%% file: neurips_2026.tex
\documentclass{article}
\usepackage[numbers,sort&compress]{natbib}
\usepackage[final,main]{neurips_2026}
\usepackage{float}

\usepackage[utf8]{inputenc} %
\usepackage[T1]{fontenc}    %
\usepackage{hyperref}       %
\usepackage{xurl}            %
\usepackage{booktabs}       %
\usepackage{amsfonts}       %
\usepackage{amsmath}
\usepackage{amssymb}
\usepackage{amsthm}
\usepackage{nicefrac}       %
\usepackage{microtype}      %
\usepackage[dvipsnames]{xcolor}
\usepackage{algorithm}
\usepackage[noend]{algpseudocode}
\usepackage{pgfplots}
\usepackage{pgfplotstable}
\usepackage[outline]{contour}
\contourlength{0.6pt}

\usepackage{wrapfig}

\pgfplotsset{compat=1.18}
\usepgfplotslibrary{colorbrewer}
\usetikzlibrary{calc,arrows.meta}

\newtheorem{lemma}{Lemma}[section]
\newtheorem{theorem}{Theorem}[section]
\newtheorem{definition}{Definition}[section]

\title{A Batched Hartigan's $k$-Means Clustering}

\author{%
  Edward Raff$^{1,2,3}$ \qquad Michael Slawinski$^1$\\
  $^1$CrowdStrike\\
  $^2$University of Maryland, Baltimore County\\
  $^3$Syracuse University
}

\begin{document}

\maketitle

\begin{abstract}
Hartigan's $k$-means algorithm has theoretical advantages over the standard Lloyd's variant, which has become synonymous with ``$k$-means''. Yet, Hartigan's algorithm is far slower in practice due to its inherently sequential calculations. We devise a Batched Hartigan's method that enables vectorized calculation by exploiting the triangle inequality to expose a set of parallel-safe updates and a smaller set of sequential updates. This makes Hartigan's method competitive with Lloyd's in execution speed for the first time. 
\end{abstract}

\section{Introduction}

The k-means algorithm has been a workhorse of machine learning for over 50 years~\cite{macqueen_methods_1967}, with the simple procedure originally proposed by Lloyd still the dominant version of the algorithm fundamentally in use continuously since its proposal~\cite{lloyd1982} and first presented in 1957. Yet, another old variant of k-means was proposed by Hartigan and Wong~\cite{hartigan_algorithm_1979} has been consistently shown to be a theoretical Pareto improvement over the original Lloyd variant. As we will review soon, Hartigan and Wong's method (or Hartigan for short) converges to a strictly superior set of minima, in fewer epochs, with monotonically decreasing objective values in each iteration of the dataset. 
We posit that Hartigan's variant of k-means has been neglected due to computational cost. Despite having theoretically fewer calculations to perform, the original algorithm is entirely sequential and thus eliminates almost all opportunities for efficient vectorized operations afforded by Basic Linear Algebra Subprograms (BLAS), the standard compiled routines for vectorized linear algebra available in every scientific computing environment~\cite{lawson_basic_1979,dongarra_algorithm_1988,dongarra_set_1990}. Our implementation relies only on BLAS, specifically GEneral Matrix Multiply (GEMM), the level-3 BLAS operation computing $C = \alpha AB + \beta C$, which achieves near-peak CPU throughput through cache blocking and single-instruction, multiple-data (SIMD) vectorization. We use GEMM for batch distance computation via the identity $\|\mathbf{x}-\mathbf{c}\|^2 = \|\mathbf{x}\|^2 - 2\mathbf{x}\cdot\mathbf{c} + \|\mathbf{c}\|^2$. Production BLAS uses the classical matrix-multiplication complexity: sub-cubic algorithms such as Strassen have large constants and weaker numerical stability and, to our knowledge, are not used in production BLAS. The advantage here is therefore constant-factor cache and SIMD efficiency, not a different asymptotic complexity.

By using the triangle inequality, we expose parallelism in the computation of Hartigan's algorithm that was not possible previously. Two key insights are required for our approach. First, in a modern system with large datasets and large $k$, the maximum change on a per-item basis to a given centroid is small, which enables safely determining most cluster assignments in bulk operations. Second, that the random order normally used for Hartigan's algorithm is not a necessary condition for any of its desirable properties, and so bulk operations can re-order points into a ``parallel'' set and a smaller ``sequential'' set, allowing us to vectorize the implementation and significantly improve runtime. 

The rest of this article is organized as follows. First, we will review the related work to our own in \autoref{sec:related_work}, covering the theory motivating the use of Hartigan's algorithm, along with the development of triangle inequality and faster k-means variants. Second, we will describe Hartigan-Wong's variation of $k$-means in detail in \autoref{sec:background}, and its inherently sequential design. This motivates our improved variant in \autoref{sec:fast_hartigan}, where we use the triangle inequality to not just prune calculations, but expose batch calculations not previously available for efficient vectorization. Our results will show in \autoref{sec:results} that our Batched Hartigan algorithm maintains the same solution quality as the standard Hartigan algorithm, is usually $10\times$ faster than previous implementations of Hartigan's algorithm, and is competitive with modern fast implementations of Lloyd's $k$-means algorithm. Finally, we will conclude in \autoref{sec:conclusion}. 

\section{Related Work} \label{sec:related_work}

Hartigan's algorithm was originally faster on older hardware~\cite{hartigan_algorithm_1979}, but is now misaligned with modern hardware efficiencies compared to variants of Lloyd's algorithm that updated the centroids once after viewing all data points. 
Hartigan's converges to a smaller and superior set as formally shown by ~\cite{pmlr-v9-telgarsky10a}, though the benefit is smaller when using good quality initial centers (e.g., via k-means++~\cite{Arthur2007,Raff2021}). ~\cite{Slonim:2013:HKV:2540128.2540369} further refined this by generalizing Hartigan's to use any Bregman divergence and showing how it is superior at avoiding degenerate cluster solutions. 
~\cite{manthey_worst-case_2024}
However, a string of works have been steadily showing that Hartigan's method has theoretically justifiable superiority over Lloyds. 
~\cite{lederman2026} showed that Hartigan's is significantly superior in clustering in sparse and high dimensional problems. 
Others ~\cite{nielsen_further_2014,clement_effective_2026} have proposed further improvements to Hartigan's method as well. In this article, we focused on the computational aspect of Hartigan's method and make it computationally viable for users to apply to their datasets, as we find all current implementations are unreasonably slow. 

Using the triangle inequality to prune out redundant calculations has been popular since the 90s~\cite{Pelleg:1999:AEK:312129.312248}. Elkan's~\cite{elkan2003} and Hammerly's ~\cite{hamerly2010} are the most popular today that we include as comparisons for runtime evaluation. Many works have further proposed different strategies for enabling pruning and balancing distance calculations vs efficient calculation~\cite{Curtin2017,Bottesch2016,hamerly2015accelerating,Ding:2015:YKD:3045118.3045181,Phillips2002}, but an exhaustive comparison is beyond our scope. Our work shows how to use these methods not just to avoid redundant calculations, but a novel approach of enabling vectorized calculations via the triangle inequality rather than sacrificing bulk calculation to prune. 

Other work accelerated Lloyd's brute-force $k$-means algorithm have been developed using GPUs to leverage their efficacy in large batch calculations and obtain $\approx100\times$ speedups over the naive batch Lloyd on CPU~\cite{yang_flash-kmeans_2026,li_large_2023,cordeiro_parallel_2024,li_speeding_2013,lutz_efficient_2018}. However, very few have tried to use both a GPU and triangle-inequality based pruning, and results are far more modest with 3-8$\times$ speedup over the same CPU algorithm~\cite{taylor_accelerating_2021}. Better designing triangle-inequality pruning for effective GPU parallelization is still an open problem, and our results are not sufficient for GPU parallelism today --- but it shows the potential path for future work in fast GPU clustering~\cite{koutra_cuslink_2023,Nolet2020,Nolet2022}.

\begin{table}[!h]
\centering
\caption{Notation used throughout this article for the Batched Hartigan method.}
\label{tab:batched-notation}
\small
\begin{tabular}{llp{0.60\textwidth}}
\toprule
Symbol & Domain & Meaning \\
\midrule
$B$
  & $\mathbb{N}$
  & Batch size (points per GEMM) \\
$\mathcal{B}$
  & index set
  & Ordered set of $B$ selected point indices; positions are relabeled
    $b=1,\dots,B$ within the batch \\
$\boldsymbol{\mu}_j^{(0)}$
  & $\mathbb{R}^{D}$
  & Centroid of cluster $j$ at batch start \\
$\boldsymbol{\mu}_j^{(b)}$
  & $\mathbb{R}^{D}$
  & Centroid of cluster $j$ after the $b$-th item is processed \\
$n_j^{(b)}$
  & $\mathbb{N}$
  & Size of cluster $j$ after the $b$-th item \\
$\hat{d}_{bj}$
  & $\mathbb{R}_{\geq 0}$
  & Precomputed $\|\mathbf{x}_b-\boldsymbol{\mu}_j^{(0)}\|^2$ (from GEMM) \\
$\delta_j^{(b)}$
  & $\mathbb{R}_{\geq 0}$
  & Cumulative drift bound for cluster $j$ after $b$ items \\
$H_j^{\mathrm{lo}}, H_j^{\mathrm{hi}}$
  & $\mathbb{R}$
  & Lower/upper bounds on Hartigan cost~\eqref{eq:hartigan-dist}
    to cluster $j$ \\
$u_i, \ell_i$
  & $\mathbb{R}_{\geq 0}$
  & Per-point upper/lower distance bounds (cross-batch screening) \\
$\sigma_j$
  & $\mathbb{R}_{\geq 0}$
  & Cumulative centroid displacement since the last batch touching point $i$ \\
\bottomrule
\end{tabular}
\end{table}

\section{Hartigan-Wong K-Means} \label{sec:background}

\input{sections/hartigan_wong}

\section{Fast Batched Hartigan-Wong} \label{sec:fast_hartigan}
\input{sections/fast_hartigan}

\section{Results} \label{sec:results}
\input{sections/results}

\section{Conclusion} \label{sec:conclusion}
\citet{Slonim:2013:HKV:2540128.2540369} concluded ``\textit{Hartigan’s algorithm is scarcely used in practice. Perhaps it is time for a change.}'', and we elucidate 
the simple reason it did not happen: 
Hartigan's method was 
$100\times$ slower than alternatives.
Rather than using the triangle inequality to jut prune calculations, we use it to enable vectorized calculations where none existed before. 

\bibliographystyle{plainnat}
\bibliography{references}

\clearpage
\appendix

\input{sections/appendix_worked_example}
\input{sections/appendix_bounds_figure}
\input{sections/appendix_per_k_results}
\input{sections/appendix_datasets}
\input{sections/appendix_sparse_results}
\input{sections/appendix_refinement}
\input{sections/appendix_ablation}
\input{sections/appendix_proofs}

\input{sections/appendix_speedup_figure}

\newpage
\input{checklist.tex}

\end{document}

%% file: sections/hartigan_wong.tex
Let $X \in \mathbb{R}^{N \times D}$ be the input data matrix with rows
$\mathbf{x}_1, \dots, \mathbf{x}_N$.  We seek a partition of the index set
$[N] = \{1,\dots,N\}$ into $k$ non-empty clusters
$\mathcal{C} = \{C_1, \dots, C_k\}$ that minimises the $k$-means 
\begin{wrapfigure}[16]{r}{0.5\textwidth}
\vspace{-10pt}
\begin{minipage}{0.5\textwidth}
\begin{algorithm}[H]
\caption{Hartigan's $k$-Means Clustering}\label{alg:hartigan}
\begin{algorithmic}[1]
\Require Data matrix $X \in \mathbb{R}^{N \times D}$ with rows
         $\mathbf{x}_1,\dots,\mathbf{x}_N$;
         number of clusters $k$;
         initial partition $\mathcal{C} = \{C_1, \dots, C_k\}$
         (each $C_j \neq \varnothing$) and centroids
         $\{\boldsymbol{\mu}_j\}_{j=1}^{k}$.
\Repeat \textcolor{ForestGreen}{\textbf{ Repeat full data sweeps}}
  \For{$i = 1, \dots, N$}
    \State Let $m$ be the cluster index s.t. $i \in C_m$

\Statex \textcolor{ForestGreen}{\textbf{Evaluate every destination}}
      \For{$j = 1, \dots, k$}
        \State $\Delta^2_j \gets\Delta^{2}_{H}(\mathbf{x}_i,\, C_j)$ \Comment{\autoref{eq:hartigan-dist}}
      \EndFor
      \State $j^{\star} \gets \arg\min_{j}\; \Delta^2_j$
\Statex \textcolor{ForestGreen}{\textbf{Apply a beneficial move}}
      \If{$j^{\star} \neq m$ \textbf{and}
          $\Delta^2_{j^{\star}} < \Delta^2_m$} \Comment{Move $\mathbf{x}_i$: $C_m \to C_{j^{\star}}$}
        \State $C_m \gets C_m \setminus \{i\}$;\;
               $C_{j^{\star}} \gets C_{j^{\star}} \cup \{i\}$
        \State Update $\boldsymbol{\mu}_m$ and
               $\boldsymbol{\mu}_{j^{\star}}$
      \EndIf
  \EndFor
\Until{No centroids change}
\State \Return $\mathcal{C} = \{C_1, \dots, C_k\}$,\;
       $\{\boldsymbol{\mu}_1, \dots, \boldsymbol{\mu}_k\}$

\end{algorithmic}
\end{algorithm}
\end{minipage}
\end{wrapfigure}
objective
\begin{equation}\label{eq:kmeans-obj}
  \mathcal{L}(\mathcal{C})
  =
  \sum_{j=1}^{k} \sum_{i \in C_j}
    \|\mathbf{x}_i - \boldsymbol{\mu}_j\|^2
\end{equation}
where $\boldsymbol{\mu}_j
  = \frac{1}{|C_j|} \sum_{i \in C_j} \mathbf{x}_i ,$ is the centroid of cluster $C_j$.

\subsection{Hartigan Weighted Distance}

The key insight of Hartigan's method is that, when considering
whether to reassign $\mathbf{x}_i$ from its current cluster $C_m$ to a
candidate cluster $C_j$, one must account for the change in cluster
sizes and the resulting shift of the centroids.  This is captured by
the \emph{Hartigan weighted distance}~\citep{hartigan1975}:
\begin{equation}\label{eq:hartigan-dist}
  \Delta^{2}_{H}(\mathbf{x}_i,\, C_j)
  =
  \begin{cases}
    \frac{|C_j|}{|C_j| - 1}\;
      \|\mathbf{x}_i - \boldsymbol{\mu}_j\|^2,
     \text{ if } i \in C_j%
    \\
    \frac{|C_j|}{|C_j| + 1}\;
      \|\mathbf{x}_i - \boldsymbol{\mu}_j\|^2,
     \text{ if } i \notin C_j%
  \end{cases}
\end{equation}
The weighting $|C_j|/(|C_j|-1)$ inflates the squared distance to the current
cluster (penalising self-influence of $\mathbf{x}_i$ on its own centroid),
while $|C_j|/(|C_j|+1)$ deflates the squared distance to a candidate cluster
(anticipating the centroid shift upon absorption of $\mathbf{x}_i$).
Algorithm~\ref{alg:hartigan} reassigns $\mathbf{x}_i$ to
$j^{\star}=\arg\min_j\Delta_H^2(\mathbf{x}_i,C_j)$ whenever
$\Delta_H^2(\mathbf{x}_i,C_{j^{\star}})
<\Delta_H^2(\mathbf{x}_i,C_m)$, the greedy move that maximally decreases
the $k$-means objective~\eqref{eq:kmeans-obj}~\citep{telgarsky2010}.

Despite its simplicity, three valuable properties have been proven using Hartigan's method:
\textbf{(1) Monotone decrease:}  Each accepted reassignment
        strictly decreases $\mathcal{L}(\mathcal{C})$, so the algorithm
        terminates in finitely many steps.
\textbf{(2) No empty clusters:}  When $N \geq k$ distinct points
        exist, the output partition has no empty
        clusters~\citep{telgarsky2010}.
\textbf{(3) Strictly finer fixed points:}  The set of local optima
        of Hartigan's method is a (possibly strict) subset of those of
        Lloyd's method~\citep{telgarsky2010}.  In high-dimensional,
        high-noise regimes, Lloyd's algorithm can become trapped at
        essentially every balanced partition, whereas Hartigan's
        algorithm provably avoids these spurious fixed
        points~\citep{lederman2026}.

%% file: sections/fast_hartigan.tex
This section introduces the batched method and its components.  For
$z \in \mathbb{R}$, write $[z]_+ \mathrel{:=} \max\{0,z\}$ for the
positive part of~$z$.  In paired assignments, the $+$ branch supplies
the first target and the $-$ branch supplies the second.
\autoref{tab:algorithm-state} summarizes the state used by
\autoref{alg:batched-hartigan} before the detailed definitions below.
\begin{table}[H]
\centering
\caption{State used by Batched Hartigan.}
\label{tab:algorithm-state}
\small
\begin{tabular}{lp{0.74\textwidth}}
\toprule
Symbol & Meaning and detailed definition \\
\midrule
$\hat{d}_{bj},\delta_j^{(b)},H_j^{\mathrm{lo}},H_j^{\mathrm{hi}}$ & Batch-start squared
distance, drift bound, and Hartigan-cost bounds; \S\ref{sec:safe-reassign} \\
$\boldsymbol{\mu}_j^{(b)},n_j^{(b)}$ & Centroid and size of cluster $j$
after batch position $b$; \S\ref{sec:safe-reassign} \\
$B,B_{\min},B_{\max}$ & Current and limiting batch sizes; \S\ref{sec:adaptive} \\
$\hat{p},\lambda,c,p_{\mathcal B},n_{\mathrm{moved}}$ & Exponential moving
average (EMA) state and batch-size inputs; \S\ref{sec:adaptive} \\
$\mathcal{B},b$ & Ordered selected point indices \& their relabeled
batch positions $b=1,\dots,B$; \S\ref{sec:pruning} \\
$m(i)$ & Cluster currently assigned to point $i$; \S\ref{sec:pruning} \\
$u_i,\ell_i,\sigma_i^u,\Sigma_i^\ell,\sigma_j,\Sigma_{\max}$ &
Cross-batch distance bounds and lazy drift state; \S\ref{sec:pruning} \\
\bottomrule
\end{tabular}
\end{table}

Standard Hartigan (Algorithm~\ref{alg:hartigan}) is inherently
sequential: each reassignment updates two centroids, and the next
point's distances are evaluated against the updated state.  This
prevents the use of level-3 BLAS routines for the distance
computation.  Empirically, for $k \geq 32$ and/or $D \geq 50$, the
per-iteration cost of a Hartigan sweep exceeds that of a Lloyd
E-step by a factor of $3$--$10\times$, even 
\begin{wrapfigure}[34]{r}{0.65\textwidth}
\vspace{-15pt}
\begin{minipage}{0.65\textwidth}
\begin{algorithm}[H]
\caption{Batched Hartigan $k$-Means}\label{alg:batched-hartigan}
\small
\begin{algorithmic}[1]

\Require Data matrix $X \in \mathbb{R}^{N \times D}$;
         number of clusters $k$;
         smoothing parameter $\lambda \in (0,1)$;
         safety factor $c \in (0,1)$;
         bounds $B_{\min}, B_{\max}$;
         initial partition $\mathcal{C}$
         (each $C_j \neq \varnothing$).
\State Initialise centroids $\boldsymbol{\mu}_j$,
       sizes $n_j$ for $j=1,\dots,k$
       \Comment{\textcolor{ForestGreen}{\S\ref{sec:safe-reassign}}}
\State Precompute
       $\|\mathbf{x}_i\|^{2}$ for all $i \in [N]$
       \Comment{\textcolor{ForestGreen}{reused across all batches}}
\State $\hat{p} \gets 1$
       \Comment{\textcolor{ForestGreen}{%
         exponential moving average; \S\ref{sec:adaptive}}}
\State $B \gets B_{\max}$
       \Comment{\textcolor{ForestGreen}{\S\ref{sec:adaptive}}}
\State Initialise per-point bounds
       $u_i \gets \infty,\;
       \ell_i \gets 0,\;
       \sigma_i^{u} \gets 0,\;
       \Sigma_i^{\ell} \gets 0$
       for all $i$
       \Comment{\textcolor{ForestGreen}{%
         \S\ref{sec:pruning}}}
\State $\sigma_j \gets 0$ for $j = 1,\dots,k$
\State $\Sigma_{\max} \gets 0$
       \Comment{\textcolor{ForestGreen}{\S\ref{sec:pruning}}}

\Repeat \Comment{\textcolor{ForestGreen}{\textbf{Process epochs and batches}}}
  \For{\textbf{each batch} $\mathcal{B}$
       \textbf{from}
       \Call{BuildBatch}{$B$}}
       \Comment{\textcolor{ForestGreen}{%
         Algorithm~\ref{alg:build-batch}}}
    \State $\boldsymbol{\mu}_j^{(0)} \gets \boldsymbol{\mu}_j$ and
           $n_j^{(0)} \gets n_j$ for all $j$
           \Comment{\textcolor{ForestGreen}{batch-start state}}
    \State $\hat{d}_{bj}
           \gets \|\mathbf{x}_b\|^{2}
                 - 2\,\mathbf{x}_b
                      \cdot \boldsymbol{\mu}_j^{(0)}
                 + \|\boldsymbol{\mu}_j^{(0)}\|^{2}$
           \;\textbf{for all}\; $b \in \mathcal{B},\; j \in [k]$
    \label{line:gemm}
    \State $(\{\delta_j^{(|\mathcal{B}|)}\},\, n_{\mathrm{moved}})
           \gets
           \Call{PartialParallelBatch}{\mathcal{B},\,
                                \{\hat{d}_{bj}\},\,
                                \{\boldsymbol{\mu}_j^{(0)}\},\,
                                \{n_j^{(0)}\}}$
           \Comment{\textcolor{ForestGreen}{%
             Alg.~\ref{alg:partial-parallel-batch}}}
\Statex \textcolor{ForestGreen}{\textbf{Refresh cross-batch bounds}}
    \State $\sigma_j \gets
           \sigma_j + \delta_j^{(|\mathcal{B}|)}$
           \;\textbf{for each}\; $j$
           \Comment{\textcolor{ForestGreen}{%
             refresh bounds (\S\ref{sec:pruning})}}
    \State $\Sigma_{\max} \gets \Sigma_{\max}
             + \max_j \delta_j^{(|\mathcal{B}|)}$
    \For{batch position $b$ corresponding to dataset index $i$}
      \State $u_i \gets \sqrt{\hat{d}_{b,\,m(i)}}
             + \delta_{m(i)}^{(|\mathcal{B}|)}$
      \State $\sigma_i^{u} \gets \sigma_{m(i)}$
      \State $\ell_i \gets
             \bigl[\min_{j \neq m(i)}
             \sqrt{\hat{d}_{bj}}
             - \max_j \delta_j^{(|\mathcal{B}|)}\bigr]_+$
      \State $\Sigma_i^{\ell} \gets \Sigma_{\max}$
    \EndFor
\Statex \textcolor{ForestGreen}{\textbf{Select the next batch size}}
    \State $\boldsymbol{\mu}_j
           \gets \boldsymbol{\mu}_j^{(|\mathcal{B}|)}$ and
           $n_j \gets n_j^{(|\mathcal{B}|)}$ for all $j$
    \State $p_{\mathcal{B}} \gets
           n_{\mathrm{moved}} / |\mathcal{B}|$
           \Comment{\textcolor{ForestGreen}{\S\ref{sec:adaptive}}}
    \State $B \gets
           \Call{DetermineBatchSize}{\{n_j^{(|\mathcal{B}|)}\},\,
                                     p_{\mathcal{B}},\,
                                     \hat{p},\, \lambda,\, c}$
           \Comment{\textcolor{ForestGreen}{%
             Algorithm~\ref{alg:batch-size}}}
  \EndFor
\Until{No centroids change}

\State \Return $\mathcal{C},\;
       \{\boldsymbol{\mu}_1,\dots,\boldsymbol{\mu}_k\}$

\end{algorithmic}
\end{algorithm}
\end{minipage}
\end{wrapfigure}
though both perform
$\Theta(NkD)$ distance evaluations, because Lloyd's distances
are computed by a single highly-optimized matrix product
(\autoref{tab:impl-speedup}).

We propose \emph{Batched Hartigan}, a framework that groups
$B$ consecutive points into a batch, precomputes all $B \times k$
distances in one GEMM, and then processes the batch---using the
triangle inequality to certify that precomputed distances remain
valid despite ongoing centroid drift.
The subsections that follow develop each component in turn:
safe sequential processing~(\S\ref{sec:safe-reassign}),
parallel batch execution~(\S\ref{sec:parallel}),
adaptive batch sizing~(\S\ref{sec:adaptive}),
and cross-batch pruning of stable points~(\S\ref{sec:pruning}).
Algorithm~\ref{alg:batched-hartigan} provides the high-level
structure.

Throughout this section, the notation in \autoref{tab:batched-notation}
distinguishes quantities fixed at batch start from quantities updated after
each item.

To re-assign points to new centroids/clusters, we must bound the distance between a centroid and it's clusters potential locations $\mu^{(b)}$. Because Hartigan weights the update by cluster size, this value is meaningfully bounded. To demonstrate, we must start by defining the drift in the clusters:

\subsection{Safe Reassignment}\label{sec:safe-reassign}
\leavevmode\par\vspace{-\baselineskip}
\par\addvspace{\topsep}
\noindent\refstepcounter{definition}\label{def:cum-drift}%
\textbf{Definition~\thedefinition} (Cumulative drift bound).
{\itshape
Let a batch begin with centroids
$\boldsymbol{\mu}_j^{(0)}$, and let
$\boldsymbol{\mu}_j^{(1)}, \boldsymbol{\mu}_j^{(2)}, \dots$
denote the centroid of cluster~$j$ after each successive
reassignment within the batch.  The cumulative drift bound is
$\delta_j^{(b)}
= \sum_{s=1}^{b}
  \|\boldsymbol{\mu}_j^{(s)}
         - \boldsymbol{\mu}_j^{(s-1)}\|$.
By the triangle inequality,
$\|\boldsymbol{\mu}_j^{(b)} - \boldsymbol{\mu}_j^{(0)}\|
 \leq \delta_j^{(b)}$.
If cluster $j$ is not involved in any reassignment during
steps $1,\dots,b$, then $\delta_j^{(b)} = 0$.
\par}

This then allows us to define upper and lower-bounds on the Hartigan cost \autoref{eq:hartigan-dist}. For brevity, paired signs use subtraction for the lower bound and addition for the upper bound.  Since both $\sqrt{\hat{d}_{bj}}$ and $\delta_j^{(b-1)}$ are nonnegative, the positive-part operator is redundant in the upper bound.

\begin{lemma}[Distance bound propagation]%
\label{lem:dist-bounds}
Let $\hat{d}_{bj}
= \|\mathbf{x}_b - \boldsymbol{\mu}_j^{(0)}\|^{2}$
be the precomputed squared distance.  After $b{-}1$ items in the
batch have been processed (with cumulative drift
$\delta_j^{(b-1)}$ for cluster~$j$,
Definition~\ref{def:cum-drift}), where $m$ denotes the index of
$\mathbf{x}_b$'s currently assigned cluster, the true squared distance and
Hartigan cost satisfy: \textbf{(i) Squared-distance bounds:} By the triangle inequality applied to
$\|\mathbf{x}_b - \boldsymbol{\mu}_j^{(b-1)}\|$
and \autoref{def:cum-drift}, is bounded above/below by
$\bigl[\sqrt{\hat{d}_{bj}} \pm \delta_j^{(b-1)}\bigr]_+^{2}$.
\textbf{(ii) Hartigan cost bounds:} Since the cluster sizes $n_j^{(b-1)}$ are tracked exactly, the
Hartigan cost~\eqref{eq:hartigan-dist} is bounded above/below (defining $H^\text{hi}$ and $H^\text{lo}$):
$\Delta_H^{2}\!\bigl(\mathbf{x}_b,\, C_m^{(b-1)}\bigr)$ by
$\frac{n_m^{(b-1)}}{n_m^{(b-1)} - 1}
\bigl[\sqrt{\hat{d}_{bm}} \pm \delta_m^{(b-1)}\bigr]_+^{2}$
for the current cluster $b \in C_m$.
The candidate cluster case for $j \neq m$ is instead bounded above/below by  
$\frac{n_j^{(b-1)}}{n_j^{(b-1)} + 1}
\bigl[\sqrt{\hat{d}_{bj}} \pm \delta_j^{(b-1)}\bigr]_+^{2}$.
\end{lemma}
\begin{wrapfigure}[36]{r}{0.6\textwidth}
\vspace{-40pt}
\begin{minipage}{0.6\textwidth}
\begin{algorithm}[H]
\caption{\textsc{SequentialBatch}: per-point processing with
         bounds}\label{alg:sequential-batch}
\small
\begin{algorithmic}[1]

\Require Ordered set $\mathcal{A} \subseteq \mathcal{B}$;
         precomputed distances $\{\hat{d}_{bj}\}$;
         state
         $\{\boldsymbol{\mu}_j^{(b_0)},n_j^{(b_0)},
         \delta_j^{(b_0)}\}$ after $b_0$ decisions.
\Statex \textcolor{ForestGreen}{\textbf{Initialize within-batch state}}
\State Use $b_0=0$ and $\delta_j^{(0)}=0$ for a full sequential batch
\State $n_{\mathrm{moved}} \gets 0$
\For{$r = 1, \dots, |\mathcal{A}|$} \Comment{\textcolor{ForestGreen}{\textbf{Process points in order}}}
  \State $b \gets b_0+r$ and let $\mathbf{x}_b$ be item $r$ of $\mathcal{A}$
  \State Set
         $(\boldsymbol{\mu}_j^{(b)},n_j^{(b)},\delta_j^{(b)})
         \gets
         (\boldsymbol{\mu}_j^{(b-1)},n_j^{(b-1)},\delta_j^{(b-1)})$
         for all $j$
  \State $m \gets$ cluster of $\mathbf{x}_b$
  \If{$n_m^{(b-1)} \leq 1$} \textbf{continue} \EndIf

\Statex \textcolor{ForestGreen}{\textbf{Bound every Hartigan cost}}
  \State $(H_m^{\mathrm{hi}},H_m^{\mathrm{lo}}) \gets
         \dfrac{n_m^{(b-1)}}{n_m^{(b-1)}-1}
         \bigl[\sqrt{\hat{d}_{bm}}
         \pm\delta_m^{(b-1)}\bigr]_+^{2}$
         \Comment{\textcolor{ForestGreen}{%
           Lemma~\ref{lem:dist-bounds}}}

  \For{$j = 1,\dots,k,\; j \neq m$}
    \State $(H_j^{\mathrm{hi}},H_j^{\mathrm{lo}}) \gets
           \dfrac{n_j^{(b-1)}}{n_j^{(b-1)}+1}
           \bigl[\sqrt{\hat{d}_{bj}}
           \pm\delta_j^{(b-1)}\bigr]_+^{2}$
  \EndFor

  \State $j^{\star} \gets \arg\min_{j \neq m}
                                   H_j^{\mathrm{lo}}$

\Statex \textcolor{ForestGreen}{\textbf{Apply S1, S2, or exact fallback}}
  \If{$H_{j^{\star}}^{\mathrm{hi}} < H_m^{\mathrm{lo}}$
      \textbf{and}
      $H_{j^{\star}}^{\mathrm{hi}} < H_j^{\mathrm{lo}}$
      for all $j \neq j^{\star}, m$}
      \Comment{\textcolor{ForestGreen}{%
        safe move (S1)}}

    \State $\delta_m^{(b)} \gets \delta_m^{(b-1)}+
           \|\mathbf{x}_b
             - \boldsymbol{\mu}_m^{(b-1)}\|
             \,/\, (n_m^{(b-1)} - 1)$
    \State $\delta_{j^{\star}}^{(b)}
           \gets \delta_{j^{\star}}^{(b-1)}+
           \|\mathbf{x}_b
             - \boldsymbol{\mu}_{j^{\star}}^{(b-1)}\|
             \,/\, (n_{j^{\star}}^{(b-1)} + 1)$
    \State Update
           $\boldsymbol{\mu}_m^{(b)},\,
            \boldsymbol{\mu}_{j^{\star}}^{(b)}$
           incrementally
    \State $n_m^{(b)} \gets n_m^{(b-1)}-1$
    \State $n_{j^{\star}}^{(b)}
           \gets n_{j^{\star}}^{(b-1)}+1$
    \State $n_{\mathrm{moved}} \gets n_{\mathrm{moved}} + 1$

  \ElsIf{$H_j^{\mathrm{lo}} \geq H_m^{\mathrm{hi}}$
         for all $j \neq m$}
         \Comment{\textcolor{ForestGreen}{%
           safe stay (S2)}}

    \State \textbf{continue}

  \Else
         \Comment{\textcolor{ForestGreen}{%
           ambiguous: recompute exact distances}}

    \State Compute exact
           $\|\mathbf{x}_b -
              \boldsymbol{\mu}_j^{(b-1)}\|^{2}$
           for all $j$
    \State Evaluate exact Hartigan costs
    \State Reassign if beneficial;\;
           update
           $\delta_m^{(b)},\delta_{j^{\star}}^{(b)},
           \boldsymbol{\mu}_m^{(b)},
           \boldsymbol{\mu}_{j^{\star}}^{(b)},
           n_m^{(b)},n_{j^{\star}}^{(b)}$,
           and increment $n_{\mathrm{moved}}$
           if a move occurs

  \EndIf

\EndFor

\State \Return
       $\{\boldsymbol{\mu}_j^{(b_0+|\mathcal{A}|)}\}$,
       $\{n_j^{(b_0+|\mathcal{A}|)}\}$,
       $\{\delta_j^{(b_0+|\mathcal{A}|)}\}$,
       and $n_{\mathrm{moved}}$

\end{algorithmic}
\end{algorithm}
\end{minipage}
\end{wrapfigure}
\textit{Proof sketch.}  Apply the triangle and reverse-triangle
inequalities to the cumulative centroid displacement, then multiply by
the positive Hartigan weights; see \autoref{app:proofs}.

With this in place, we can state the conditions under which we can safely perform Hartigan's clustering while using potentially stale centroid distances. Let $m$ denote the current cluster of $\mathbf{x}_b$.  Using
the Hartigan cost bounds from
Lemma~\ref{lem:dist-bounds}\textup{(ii)}, write
\begin{equation}\label{eq:Hbounds-shorthand}
  H_m^{\mathrm{lo}},\; H_m^{\mathrm{hi}}
  \quad\text{for the current-cluster bounds,}
  \qquad
  H_j^{\mathrm{lo}},\; H_j^{\mathrm{hi}}
  \quad\text{for each candidate } j \neq m.
\end{equation}

\begin{definition}[Safe reassignment decision]%
\label{def:safe-reassign}
Let
$j^{\star} = \arg\min_{j \neq m} H_j^{\mathrm{lo}}$.
The reassignment decision for $\mathbf{x}_b$ is
\emph{safe} if one of:
\textbf{(S1) Safe move:}
$H_{j^{\star}}^{\mathrm{hi}} < H_m^{\mathrm{lo}}$
and
$H_{j^{\star}}^{\mathrm{hi}} < H_j^{\mathrm{lo}}$
for all $j \neq j^{\star}, m$ (the upper-bound cost of $j^{\star}$ is below the lower-bound cost of staying and all other candidates).
\textbf{(S2) Safe stay:}
$H_j^{\mathrm{lo}} \geq H_m^{\mathrm{hi}}$
for all $j \neq m$ (every candidate's lower-bound cost exceeds the upper-bound cost of staying).
Otherwise the decision is \emph{ambiguous} and requires exact recomputation.
\end{definition}

In the Appendix \autoref{thm:batched-correct} we prove that this is exactly replicates Hartigan's algorithm under the same iteration order. However, we will later show a more useful result for re-ordering the items in a batch, and so deffer it's details.

Having defined the conditions of safe updates, we combine them all into \autoref{alg:sequential-batch} for safe sequential evaluation of all points, following directly from the prior Lemmas and definitions. Bounds on the Hartigan cost \autoref{eq:hartigan-dist} are calculated from an initial bulk computation, and only violating points need to be recomputed. The sequential kernel performs at most $2\times$ the arithmetic of a plain Hartigan sweep: once for the batch GEMM distances, $O(BkD)$, and at most once more for exact recomputation of ambiguous points. In practice, fewer than 5\% of points require recomputation in steady state, so this overhead is small.

\subsection{Simultaneous Execution}\label{sec:parallel}

The sequential inner loop processes $B$~points in series,
evaluating bounds one at a time.  We now classify each point's
tentative decision as \emph{safe}---certifiable under worst-case
drift bounds---or \emph{ambiguous}. 
The safe majority is executed
in a single parallel bulk update; the small ambiguous remainder
is processed sequentially afterward, with drift initialized to
the exact displacement produced by the safe execution.

\begin{definition}[Tentative decisions and aggregate flow]%
\label{def:tentative}
Given precomputed distances
$\hat{d}_{bj}
= \|\mathbf{x}_b - \boldsymbol{\mu}_j^{(0)}\|^{2}$
and pre-batch sizes $n_j$, the \emph{tentative decision}
$T(b)$ for point~$b$ is the standard Hartigan comparison
(Algorithm~\ref{alg:hartigan}, lines 4--9) evaluated with
exact distances $\hat{d}_{bj}$ and sizes $n_j$.  The
\emph{aggregate flow} for each cluster~$j$ is
$L_j = |\{b : T(b) \text{ removes } b \text{ from } C_j\}|$,
$J_j = |\{b : b \text{ joins } C_j \text{ under } T(b)\}|$.
\end{definition}

Once we have the tentative leave decisions $L$ and join decisions $J$,
we need to know that we can re-process the points in a safer order.
\autoref{lem:commutative} shows that these updates commute.

\begin{lemma}[Commutativity of centroid updates]%
\label{lem:commutative}
Let cluster~$C_j$ have centroid $\boldsymbol{\mu}_j$ and size
$n_j$.  Suppose a set of points $S^{-} \subsetneq C_j$ departs and
a disjoint set $S^{+} \subset [N] \setminus C_j$ arrives.  Then the updated centroid
$\boldsymbol{\mu}_j'
= ({n_j\,\boldsymbol{\mu}_j
      - \sum_{i \in S^{-}} \mathbf{x}_i
      + \sum_{i \in S^{+}} \mathbf{x}_i})
   / ({n_j - |S^{-}| + |S^{+}|})$
is independent of the order in which the individual removals and
insertions are applied.
\end{lemma}
\begin{wrapfigure}[34]{r}{0.65\textwidth}
\vspace{-25pt}
\begin{minipage}{0.65\textwidth}
\begin{algorithm}[H]
\caption{\textsc{PartialParallelBatch}: safe/ambiguous partition
         execution}\label{alg:partial-parallel-batch}
\small
\begin{algorithmic}[1]

\Require Batch $\mathcal{B}$;
         precomputed distances $\{\hat{d}_{bj}\}$;
         batch-start centroids $\{\boldsymbol{\mu}_j^{(0)}\}$;
         batch-start sizes $\{n_j^{(0)}\}$.
\Ensure  Updates
         $\{\boldsymbol{\mu}_j^{(|\mathcal{B}|)},
         n_j^{(|\mathcal{B}|)}\}$ and returns
         $\{\delta_j^{(|\mathcal{B}|)}\}$ and $n_{\mathrm{moved}}$.

\Statex \textcolor{ForestGreen}{\textbf{Form tentative decisions in parallel}}
\For{$b \in \mathcal{B}$ \textbf{in parallel}}
       \Comment{\textcolor{ForestGreen}{%
         tentative decisions
         (Definition~\ref{def:tentative})}}
  \State $m \gets$ cluster of $\mathbf{x}_b$
  \State $T(b) \gets$ standard Hartigan decision using
         $\hat{d}_{bj}$ and $n_j^{(0)}$
\EndFor

\Statex \textcolor{ForestGreen}{\textbf{Compute aggregate drift bounds}}
\State Compute $L_j, J_j, M_j, R_j$ for each $j$
       \Comment{\textcolor{ForestGreen}{Def.~\ref{def:tentative},
       Lemma~\ref{lem:aggregate-bounds}}}
\If{$M_j \geq n_j^{(0)} - L_j$ for any $j$}
  \State \Return $\Call{SequentialBatch}{\mathcal{B},\,
         \{\hat{d}_{bj}\},\,
         \{\boldsymbol{\mu}_j^{(0)},n_j^{(0)},
         \delta_j^{(0)}=0\},\, b_0=0}$
       \Comment{\textcolor{ForestGreen}{%
         validity fails $\to$ full sequential}}
\EndIf
\State $\delta_j^{\mathrm{agg}} \gets
       R_j / (n_j^{(0)} - L_j - M_j)$
       for each $j$
       \Comment{\textcolor{ForestGreen}{Lemma~\ref{lem:aggregate-bounds}}}

\Statex \textcolor{ForestGreen}{\textbf{Partition safe and ambiguous points}}
\For{$b \in \mathcal{B}$ \textbf{in parallel}}
       \Comment{\textcolor{ForestGreen}{%
         per-point safety classification}}
  \State Compute worst-case Hartigan cost bounds using
         $\delta_j^{\mathrm{agg}}$
         \Comment{\textcolor{ForestGreen}{Lemma~\ref{lem:aggregate-bounds}}}
  \State $\textit{safe}[b] \gets T(b)$ satisfies (S1) or (S2)
\EndFor

\State $\mathcal{S} \gets \{b : \textit{safe}[b]\}$,\;
       $\mathcal{A} \gets \mathcal{B} \setminus \mathcal{S}$

\Statex \textcolor{ForestGreen}{\textbf{Execute the resulting partition}}
\If{$\mathcal{A} = \varnothing$}
       \Comment{\textcolor{ForestGreen}{%
         fast path: all safe}}
  \State Execute bulk centroid
         update (Lemma~\ref{lem:commutative}) for all movers in
         $\mathcal{S}$
  \State \Return drift from bulk update
\ElsIf{$\mathcal{S} = \varnothing$}
       \Comment{\textcolor{ForestGreen}{%
         fallback: all ambiguous}}
  \State \Return $\Call{SequentialBatch}{\mathcal{B},\,
         \{\hat{d}_{bj}\},\,
         \{\boldsymbol{\mu}_j^{(0)},n_j^{(0)},
         \delta_j^{(0)}=0\},\, b_0=0}$
\Else
       \Comment{\textcolor{ForestGreen}{%
         mixed: safe first, then ambiguous}}
  \State Execute bulk centroid update for movers in
         $\mathcal{S}$
         \Comment{\textcolor{ForestGreen}{%
           Lemma~\ref{lem:commutative}}}
  \State Let
         $\{\boldsymbol{\mu}_j^{(|\mathcal{S}|)},
         n_j^{(|\mathcal{S}|)}\}$
         be the resulting state
  \State $\delta_j^{(|\mathcal{S}|)} \gets
         \|\boldsymbol{\mu}_j^{(|\mathcal{S}|)}
           - \boldsymbol{\mu}_j^{(0)}\|$
         \;\textbf{for all}\; $j$
         \Comment{\textcolor{ForestGreen}{%
           actual drift, $O(kD)$}}
  \State $\Call{SequentialBatch}{\mathcal{A},\,
         \{\hat{d}_{bj}\},\,
         \{\boldsymbol{\mu}_j^{(|\mathcal{S}|)},
         n_j^{(|\mathcal{S}|)},
         \delta_j^{(|\mathcal{S}|)}\},\,
         b_0=|\mathcal{S}|}$
         \Comment{\textcolor{ForestGreen}{%
           Alg.~\ref{alg:sequential-batch}}}
  \State \Return total per-cluster drift
\EndIf

\end{algorithmic}
\end{algorithm}
\end{minipage}
\end{wrapfigure}
\textit{Proof sketch.}  The final centroid is the final member sum
divided by the final member count, both independent of update order;
see \autoref{app:proofs}.

Given commutativity (Lemma~\ref{lem:commutative}), the final
cluster state after executing a fixed set of moves is unique
regardless of processing order.  It remains to verify that every
point's decision is also order-independent: does every point make
the same reassignment choice whether it is processed first, last,
or simultaneously with all others?  To answer this we must bound
the centroid drift and size variation that any one point might
experience at an arbitrary intermediate step via \autoref{lem:aggregate-bounds}.

\begin{lemma}[Aggregate batch bounds]%
\label{lem:aggregate-bounds}
Let $\mathcal{B}$ be a batch with tentative decisions
$\{T(b)\}_{b=1}^{B}$ and aggregate flow $L_j, J_j$
(Definition~\ref{def:tentative}).  Write
$n_j^{\min} = n_j - L_j$,
$M_j = L_j + J_j$,
$R_j = \sum_{b:\,T(b)\text{ touches }C_j} \sqrt{\hat{d}_{bj}}$.
Here, ``$T(b)$ touches $C_j$'' means that $T(b)$ either removes $b$
from $C_j$ (a leave counted by $L_j$) or adds $b$ to $C_j$
(a join counted by $J_j$).
Assume $M_j < n_j^{\min}$ for all~$j$.
The \emph{aggregate drift bound} is
$\delta_j^{\mathrm{agg}} = R_j / (n_j^{\min} - M_j)$.
Then at every intermediate step~$s$ during sequential processing
of $\mathcal{B}$ in any order:
\textbf{(i)}~$\|\boldsymbol{\mu}_j^{(s)} - \boldsymbol{\mu}_j^{(0)}\|
 \leq \delta_j^{\mathrm{agg}}$ for all~$j$.
\textbf{(ii)}~The Hartigan cost of point~$b$ (in cluster~$m$)
is bounded above by
$w_m^{\max} (\sqrt{\hat{d}_{bm}} + \delta_m^{\mathrm{agg}})^{2}$
and below by
$w_j^{\min}
\bigl[\sqrt{\hat{d}_{bj}} - \delta_j^{\mathrm{agg}}\bigr]_+^{2}$,
where
$w_m^{\max} = (n_m - L_m^{-b})/(n_m - L_m^{-b} - 1)$
with $L_m^{-b} = L_m - \mathbf{1}[T(b) \text{ leaves } m]$, and
$w_j^{\min} = n_j^{\min}/(n_j^{\min} + 1)$.
\textbf{(iii)}~For safe-move~(S1) verification, directions reverse:
$w_m^{\min} = (n_m + J_m)/(n_m + J_m - 1)$ and
$w_{j^{\star}}^{\max} = (n_{j^{\star}} + J_{j^{\star}})/(n_{j^{\star}} + J_{j^{\star}} + 1)$.
\end{lemma}

\textit{Proof sketch.}  Repeated triangle inequalities bound each
centroid displacement, while the minimum possible cluster sizes bound
the Hartigan weights; see \autoref{app:proofs}.

Given this, we can develop a stronger theoretical statement that the items within a batch do not need to be processed in the original order. It is possible to re-order the items within a batch, based on their prunability. This allows us to re-order them for maximal compute efficacy, while still converging to a Hartigan fixed point solution:

\begin{theorem}[Partial parallel batch correctness]%
\label{thm:partial-parallel}
Given a GEMM batch $\mathcal{B}$ with tentative decisions
$\{T(b)\}_{b \in \mathcal{B}}$, aggregate flow $L_j, J_j$
satisfying $M_j < n_j^{\min}$ for all~$j$,
and aggregate drift bound $\delta_j^{\mathrm{agg}}$
(Lemma~\ref{lem:aggregate-bounds}), partition the batch into
$\mathcal{S} = \{b \in \mathcal{B} : T(b) \text{ satisfies (S1) or (S2) under } \delta_j^{\mathrm{agg}}\}$ and
$\mathcal{A} = \mathcal{B} \setminus \mathcal{S}$.
Then:
\textbf{(a)}~The bulk centroid update
restricted to movers in $\mathcal{S}$ is correct:
every move decreases $\mathcal{L}(\mathcal{C})$.
\textbf{(b)}~Sequential processing of~$\mathcal{A}$ with drift
initialised to
$\Delta_j = \|\boldsymbol{\mu}_j^{(\text{after } \mathcal{S})}
              - \boldsymbol{\mu}_j^{(\mathrm{GEMM})}\|$
is correct: each decision uses valid Hartigan cost bounds.
\textbf{(c)}~Every accepted move strictly decreases
$\mathcal{L}(\mathcal{C})$; no cluster is emptied.  The algorithm
terminates at a Hartigan fixed point---the same characterisation
as standard Hartigan (Algorithm~\ref{alg:hartigan}).
\end{theorem}

Algorithm~\ref{alg:batched-hartigan} calls
\textsc{PartialParallelBatch}
(Algorithm~\ref{alg:partial-parallel-batch}) for each GEMM batch.
The per-point safety classification is $O(Bk)$, and centroid updates
cost $O(BD)$ per batch and $O(ND)$ across all batches in one epoch.
Both are negligible compared to the $O(BkD)$ GEMM.  In the common case
the vast majority of points are safe, and only a small
ambiguous remainder requires sequential processing.  In the
limiting case where all points are safe
($\mathcal{A} = \varnothing$), the algorithm reduces to a single
bulk update per batch; when all points are ambiguous
($\mathcal{S} = \varnothing$), it degrades gracefully to the
sequential fallback.

At convergence, the algorithm arrives at a Hartigan
fixed point (Theorem~\ref{thm:partial-parallel}(c)), which
inherits the guarantees of~\citep{telgarsky2010}:
Hartigan fixed points are a strict subset of Lloyd fixed points.
The balanced-partition traps identified
by~\citep{lederman2026} cannot arise.
The $\mathcal{S}$-then-$\mathcal{A}$ reordering may yield
different fixed points from the same initialisation, but
$\mathcal{L}$ decreases monotonically and every accepted
move is individually optimal at the time it is made.
\subsection{Adaptive Batch Size}\label{sec:adaptive}

\begin{wrapfigure}[9]{r}{0.5\textwidth}
\vspace{-65pt}
\begin{minipage}{0.5\textwidth}
\begin{algorithm}[H]
\caption{\textsc{DetermineBatchSize}: EMA-adaptive batch
         sizing}\label{alg:batch-size}
\small
\begin{algorithmic}[1]

\Require Current sizes $\{n_j^{(|\mathcal{B}|)}\}$;
         observed reassignment rate $p_{\mathcal{B}}$;
         EMA state $\hat{p}$;
         smoothing parameter $\lambda$;
         safety factor $c$;
         bounds $B_{\min}, B_{\max}$.
\Ensure  Updated $\hat{p}$ and new batch size $B$.
\Statex \textcolor{ForestGreen}{\textbf{Update the reassignment-rate estimate}}
\State $\hat{p} \gets
       (1 - \lambda)\,\hat{p}
       + \lambda\, p_{\mathcal{B}}$
       \Comment{\textcolor{ForestGreen}{%
         EMA update (\S\ref{sec:adaptive})}}

\Statex \textcolor{ForestGreen}{\textbf{Apply the drift-validity constraint}}
\State $n_{\min} \gets \min_j\, n_j^{(|\mathcal{B}|)}$
\State \Return
       $\mathrm{clamp}\!\bigl(
         \lfloor c\, k\, n_{\min}
                 / (2\,\hat{p}) \rfloor,\;
         B_{\min},\; B_{\max}\bigr)$
       \Comment{\textcolor{ForestGreen}{%
         batch size (\S\ref{sec:adaptive})}}
\end{algorithmic}
\end{algorithm}
\end{minipage}
\end{wrapfigure}
The batch size $B$ controls the GEMM shape $(B, D) \times (D, k)$
and determines how much centroid drift accumulates before distances
are refreshed.  Larger $B$ amortizes BLAS overhead better, but the
aggregate drift bound
$\delta_j^{\mathrm{agg}} = R_j / (n_j^{\min} - M_j)$
(Lemma~\ref{lem:aggregate-bounds}) grows with $B$, loosening the
safety bounds and increasing fallback to exact computation.
The validity condition $M_j < n_j^{\min}$ of
Lemma~\ref{lem:aggregate-bounds} imposes a hard upper limit on the
number of moves per batch. We thus use the constraints to inform the value of $B$ based on the frequency of updates $p$, and then use a moving average to smoothly update our estimate of $p$ over time. 

\begin{wrapfigure}[18]{r}{0.50\textwidth}
\vspace{-45pt}
\begin{minipage}{0.50\textwidth}
\begin{algorithm}[H]
\caption{\textsc{BuildBatch}: screen stable points and collect
         unsettled indices}\label{alg:build-batch}
\small
\begin{algorithmic}[1]

\Require Batch size $B$;
         per-point bounds $u_i, \ell_i, \sigma_i^{u},
         \Sigma_i^{\ell}$;
         per-cluster drift $\sigma_j$;
         global drift $\Sigma_{\max}$;
         cluster sizes $n_j$.
\Ensure  Yields batches $\mathcal{B}_1, \mathcal{B}_2, \dots$
         of unsettled point indices, each of size $\leq B$.

\State $\mathcal{B} \gets \varnothing$ \Comment{\textcolor{ForestGreen}{\textbf{Initialize an empty batch}}}
\Statex \textcolor{ForestGreen}{\textbf{Scan and screen every point}}
\For{$i = 1, \dots, N$}
  \State $u_i^{\mathrm{eff}} \gets
         u_i + (\sigma_{m(i)} - \sigma_i^{u})$
         \Comment{\textcolor{ForestGreen}{%
            bounds from (\S\ref{sec:pruning})}}
  \State $\ell_i^{\mathrm{eff}} \gets
         \ell_i - (\Sigma_{\max} - \Sigma_i^{\ell})$

  \If{$\dfrac{n_j^{\min}}{n_j^{\min}+1}
       (\ell_i^{\mathrm{eff}})^{2}
       \;\geq\;
       \dfrac{n_{m(i)}}{n_{m(i)}-1}
       (u_i^{\mathrm{eff}})^{2}$}
    \State \textbf{continue}
           \Comment{\textcolor{ForestGreen}{%
             screened --- safe stay (\S\ref{sec:pruning})}}
  \EndIf

  \State $\mathcal{B} \gets \mathcal{B} \cup \{i\}$

  \If{$|\mathcal{B}| = B$}
    \State \textbf{yield} $\mathcal{B}$
    \State $\mathcal{B} \gets \varnothing$
  \EndIf
\EndFor
\If{$\mathcal{B} \neq \varnothing$} \Comment{\textcolor{ForestGreen}{\textbf{Emit the final partial batch}}}
  \State \textbf{yield} $\mathcal{B}$
         \Comment{\textcolor{ForestGreen}{%
           remaining partial batch}}
\EndIf

\end{algorithmic}
\end{algorithm}
\end{minipage}
\end{wrapfigure}
\paragraph{Constraint on $B$.}
If the reassignment rate is $p$ (fraction of points that change cluster), a batch of $B$ points produces $\approx pB$ moves, each touching two clusters, so the expected per-cluster load is $\mathbb{E}[M_j] \approx 2pB/k$.  The validity condition $M_j < n_j^{\min}$ then requires $B < k\,n_{\min}/(2p)$ where $n_{\min} = \min_j n_j$.

\paragraph{Batch-adaptive EMA.}
We maintain an exponential moving average (EMA) of the reassignment rate, updated after every batch at $O(1)$ cost.  Let $p_{\mathcal{B}} = n_{\mathrm{moved}}/|\mathcal{B}|$ be the observed rate; the EMA update is $\hat{p} \gets (1-\lambda)\hat{p} + \lambda\,p_{\mathcal{B}}$ with $\lambda \in (0,1)$ and $\hat{p}_0 = 1$ (pessimistic).  The next batch size is $B \gets \mathrm{clamp}(\lfloor c\,k\,n_{\min}/(2\hat{p})\rfloor,\, B_{\min},\, B_{\max})$ where $c \in (0,1)$ is a safety factor (default $0.5$), $B_{\min}$ is the minimum useful GEMM size, and $B_{\max}$ is a cache-aware ceiling.  Below $B \approx 64$, fixed allocation, dispatch, and thread-synchronization costs can exceed the arithmetic time.  Fixing $B=64$ produces $1.4$--$9\times$ slowdown in \autoref{tab:ablation-components}; the practical $64$--$256$ range balances GEMM efficiency against drift accumulation.  Initially $B = \mathrm{clamp}(\lfloor k n_{\min}/4\rfloor, B_{\min}, B_{\max})$; as $\hat{p}$ decreases, $B$ grows until saturating at $B_{\max}$ in later epochs.

\subsection{Pruning Stable Points}\label{sec:pruning}

Our final insight is that some points may be so stable as to make any distance calculations at all irrelevant. In later epochs most points are stable.
By maintaining cheap per-point distance bounds across batches, we
\emph{screen} stable points before batch formation and exclude
them from the GEMM entirely (following the Hamerly
design~\citep{hamerly2010}).
For each point~$i$, $m(i)$ denotes its currently assigned cluster.  We maintain
$u_i \geq \|\mathbf{x}_i - \boldsymbol{\mu}_{m(i)}\|$ (upper bound to assigned centre) and
$\ell_i \leq \min_{j \neq m(i)} \|\mathbf{x}_i - \boldsymbol{\mu}_j\|$ (lower bound to nearest other centre), initialised to $u_i = \infty$, $\ell_i = 0$.

To avoid $O(N)$ bound updates per batch, we develop \textit{lazy} drift tracking to maintain
per-cluster cumulative drift
$\sigma_j = \sum_{t} \delta_j^{(t)}$ and a global accumulator
$\Sigma_{\max} = \sum_{t} \max_j \delta_j^{(t)}$.
When point~$i$'s bounds are set after a GEMM, we record
$\sigma_i^{u} = \sigma_{m(i)}$ and
$\Sigma_i^{\ell} = \Sigma_{\max}$.  The \emph{effective bounds}
at any later time are
$u_i^{\mathrm{eff}} = u_i + (\sigma_{m(i)} - \sigma_i^{u})$
and
$\ell_i^{\mathrm{eff}} = \ell_i - (\Sigma_{\max} - \Sigma_i^{\ell})$,
each costing $O(1)$, which is proven by \autoref{lem:cross-batch}.

\begin{lemma}[Cross-batch bound validity]%
\label{lem:cross-batch}
After any sequence of batches,
$u_i^{\mathrm{eff}} \geq \|\mathbf{x}_i - \boldsymbol{\mu}_{m(i)}\|$ and
$\ell_i^{\mathrm{eff}} \leq \min_{j \neq m(i)} \|\mathbf{x}_i - \boldsymbol{\mu}_j\|$.
\end{lemma}

\textit{Proof sketch.}  Cumulative cluster drift preserves the upper
bound, while the sum of per-batch maximum drifts preserves the lower
bound; see \autoref{app:proofs}.

Point~$i$ (in cluster~$m$) is \emph{screened} if:
$\tfrac{n_j^{\min}}{n_j^{\min} + 1}
(\ell_i^{\mathrm{eff}})^{2}
\geq
\tfrac{n_m}{n_m - 1}
(u_i^{\mathrm{eff}})^{2}$,
where $n_j^{\min} = \min_{j \neq m} n_j$.
This certifies (S2), preserving the guarantee of Theorem~\ref{thm:batched-correct}.

However, the lazy updates can make the bound excessively ``stale'', and must be refreshed. This refresh happens at two levels. After each batch with drifts $\delta_j$:
(1)~\emph{Global} ($O(k)$):
$\sigma_j \mathrel{+}= \delta_j$,
$\Sigma_{\max} \mathrel{+}= \max_j \delta_j$.
(2)~\emph{Per-point} ($O(|\mathcal{B}|)$): for each $i \in \mathcal{B}$,
$u_i \gets \sqrt{\hat{d}_{i,m(i)}} + \delta_{m(i)}$,
$\sigma_i^{u} \gets \sigma_{m(i)}$,
$\ell_i \gets
\bigl[\min_{j \neq m(i)} \sqrt{\hat{d}_{ij}}
- \max_j \delta_j\bigr]_+$,
$\Sigma_i^{\ell} \gets \Sigma_{\max}$.
All bounds carry over unchanged across epoch boundaries (warm-start).

%% file: sections/results.tex
\pgfplotsset{
  results/.style={
    every axis/.append style={font=\small},
    legend style={font=\footnotesize},
    label style={font=\small},
    tick label style={font=\footnotesize},
  }
}

All experiments use $k$-means++ initialization, $k \in \{2,4,8,16,32,64,128\}$
(capped at $\lfloor\sqrt{N}\rfloor$), and report the median
over 10 independent trials.  Within each trial, all algorithms receive
identical externally selected $k$-means++ initial centers.  All runtimes are single-threaded
wall-clock seconds on a Apple M4 Pro. Our results will show that: \textbf{(1)} Batched Hartigan maintains desirable empirical properties relative to the naive variant. \textbf{(2)} Our method is faster than the prior implementations of Hartigan's algorithm. (3) That our method is comparable to popular fast Lloyd's $k$-means variants, making it viable for practitioners to explore, and validates our strategy of creating vectorizable computations via the triangle inequality. 

Five algorithms will be compared in total: sequential
Hartigan (Python with Numba JIT; Algorithm~\ref{alg:hartigan}) and
our Batched Hartigan (Algorithm~\ref{alg:batched-hartigan}).  The
Lloyd-family baselines are standard Lloyd (Python/NumPy)~\citep{lloyd1982},
Elkan (Cython via scikit-learn)~\citep{elkan2003}, and Hamerly
(C++)~\cite{hamerly2010}. These will be evaluated on 12 datasets spanning
$N \in [11\,055,\; 5\,000\,000]$ and
$D \in [3,\; 780]$. Dataset dimensions and preprocessing are reported
in \autoref{app:datasets}.

\input{sections/results_efficacy}

\input{sections/results_implspeed}

\input{sections/results_runtime}

%% file: sections/results_efficacy.tex
\subsection{Fast Batched Hartigan Efficacy}\label{sec:efficacy}

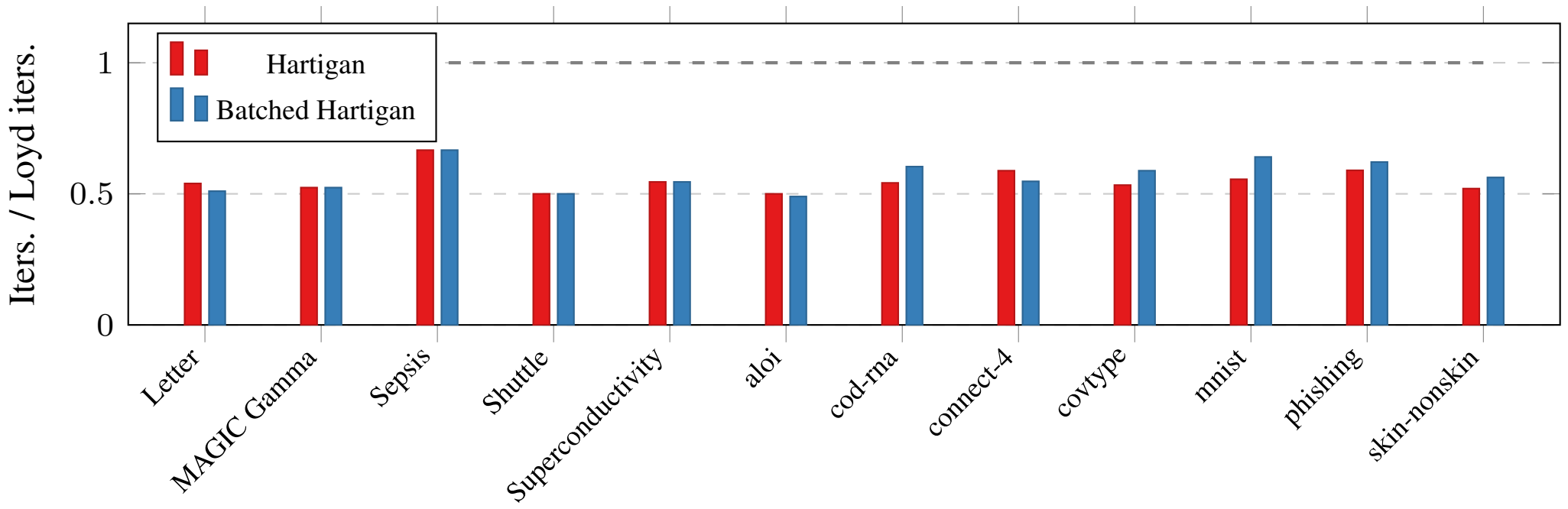
\begin{figure}[!h]
\centering
\begin{tikzpicture}
\begin{axis}[
  results,
  width=\columnwidth,
  height=0.30\columnwidth,
  ybar,
  bar width=4pt,
  symbolic x coords={
    {Letter},
    {MAGIC Gamma},
    {Sepsis},
    {Shuttle},
    {Superconductivity},
    {aloi},
    {cod-rna},
    {connect-4},
    {covtype},
    {mnist},
    {phishing},
    {skin-nonskin}
  },
  xtick=data,
  x tick label style={rotate=45, anchor=east, font=\scriptsize},
  ylabel={Iters. / Loyd iters.},
  ymin=0, ymax=1.15,
  enlarge x limits=0.06,
  legend style={at={(0.02,0.97)}, anchor=north west, font=\scriptsize},
  ymajorgrids=true,
  grid style={dashed, gray!40},
]

\addplot[fill=Set1-A, draw=Set1-A!80!black] coordinates {
  ({Letter}, 0.5395)
  ({MAGIC Gamma}, 0.5238)
  ({Sepsis}, 0.6667)
  ({Shuttle}, 0.5000)
  ({Superconductivity}, 0.5455)
  ({aloi}, 0.5000)
  ({cod-rna}, 0.5417)
  ({connect-4}, 0.5882)
  ({covtype}, 0.5333)
  ({mnist}, 0.5556)
  ({phishing}, 0.5896)
  ({skin-nonskin}, 0.5200)
};

\addplot[fill=Set1-B, draw=Set1-B!80!black] coordinates {
  ({Letter}, 0.5102)
  ({MAGIC Gamma}, 0.5238)
  ({Sepsis}, 0.6667)
  ({Shuttle}, 0.5000)
  ({Superconductivity}, 0.5455)
  ({aloi}, 0.4898)
  ({cod-rna}, 0.6038)
  ({connect-4}, 0.5476)
  ({covtype}, 0.5882)
  ({mnist}, 0.6406)
  ({phishing}, 0.6213)
  ({skin-nonskin}, 0.5625)
};

\addlegendentry{Hartigan}
\addlegendentry{Batched Hartigan}
\draw[dashed, gray, thick]
  (axis cs:{Letter},1.0) -- (axis cs:{skin-nonskin},1.0);

\end{axis}
\end{tikzpicture}
\caption{Median iteration ratio relative to Lloyd across
  $k$ values.  Both Hartigan variants converge in roughly half the
  epochs. This shows one advantage of Hartigan's method, and that our Batched variant maintains it in practice.}%
\label{fig:iter-ratio}
\end{figure}

First we validate that our Batched variant maintains desirable properties of Hartigan's method. Solution quality is measured by the sum of squared errors (SSE), the $k$-means objective $\sum_i\|\mathbf{x}_i-\boldsymbol{\mu}_{a_i}\|^2$ from \eqref{eq:kmeans-obj}, where $a_i$ denotes point~$i$'s assigned cluster. This starts with the convergence speed.
Figure~\ref{fig:iter-ratio} shows the ratio of each algorithm's
iteration count to Lloyd's $k$-means.  Both Hartigan variants
converge in roughly half the epochs: the median
Hartigan/Lloyd iteration ratio is $0.54$, and the median
Batched Hartigan/Lloyd ratio is $0.56$. 

\begin{wraptable}[15]{r}{0.55\textwidth}
\vspace{-20pt}
\centering
\caption{Median SSE ratio to Elkan across $k$ values.
  Ratio~$< 1$: Hartigan achieves lower SSE than Elkan.}\label{tab:sse-ratio}
\small
\pgfplotstableread[col sep=comma]{figures/sse_ratio_agg.csv}\sseratio
\pgfplotstabletypeset[
  columns={dataset,hartigan_ratio,batched_ratio},
  columns/dataset/.style={
    column name={Dataset},
    column type={l},
    string type,
  },
  columns/hartigan_ratio/.style={
    column name={Hartigan / Elkan},
    fixed, fixed zerofill,
    precision=4,
  },
  columns/batched_ratio/.style={
    column name={Batched / Elkan},
    fixed, fixed zerofill,
    precision=4,
  },
  every head row/.style={before row=\toprule, after row=\midrule},
  every last row/.style={after row=\bottomrule},
]{\sseratio}
\end{wraptable}
Hartigran's method converges to a superior solution set, though \cite{pmlr-v9-telgarsky10a} noted the benefit was smaller when using $k$-means++ style seeding. Still, we confirm this in our work adn validate the same quality of solution that is generally equal-or-better using our Batched Hartigan. 
Table~\ref{tab:sse-ratio} reports the ratio of each Hartigan
variant's converged SSE to Elkan's.  Values below~$1$ indicate
lower (better) SSE of Hartigan.  Both Hartigan and Batched Hartigan
consistently match or improve on Elkan's SSE, with median ratios
of $0.9999$ and $0.9999$ respectively.  This confirms that
Hartigan fixed points yield equal or better clusterings than
Lloyd fixed points (the latter being shared by Elkan and Lloyd),
and that Batched Hartigan preserves this advantage.

High-dimensional and sparse settings can produce larger differences
between Lloyd and Hartigan fixed points~\citep{lederman2026}.
\autoref{tab:sparse-quality} reports the best observed Batched
Hartigan-to-Elkan inertia ratio over the tested $k$ values and trials.

\begin{wraptable}[13]{r}{0.47\textwidth}
\vspace{-20pt}
\centering
\caption{Best observed sparse-data inertia ratio.  Ratios are Batched
Hartigan divided by Elkan; lower is better.}
\label{tab:sparse-quality}
\small
\begin{tabular}{lrrr}
\toprule
Dataset & $D$ & Inertia ratio & At $k$ \\
\midrule
breast-cancer & 10 & 0.9504 & 128 \\
leukemia & 7,129 & 0.9134 & 16 \\
duke & 7,129 & 0.8201 & 4 \\
bibtex & 1,836 & 0.9639 & 4 \\
sector & 55,197 & 0.9866 & 256 \\
news20 & 62,061 & 0.9030 & 8 \\
rcv1\_multi & 47,236 & 0.9933 & 256 \\
E2006-tfidf & 150,360 & 0.9891 & 32 \\
scotus & 126,405 & 0.9775 & 256 \\
Wiki10-31K & 104,374 & 0.9845 & 256 \\
real-sim & 20,958 & 0.9794 & 256 \\
\bottomrule
\end{tabular}
\end{wraptable}

Every dataset had at least one lower-inertia configuration; ratios
ranged from 0.8201 to 0.9933, or 0.7--18.0\% lower inertia.  Improvement
is not expected at every $k$: very small or large $k$ can yield simpler
optima for which Lloyd and Hartigan solutions coincide.  Density and
runtime results are reported in \autoref{app:sparse-results}.

%% file: sections/results_implspeed.tex
\subsection{Implementation Speed}\label{sec:impl-speed}

\begin{wraptable}[13]{r}{0.52\textwidth}
\vspace{-24pt}
\centering
\caption{Median wall time across $k$ values for three Hartigan
  implementations, in seconds.}\label{tab:impl-speedup}
\small
\setlength{\tabcolsep}{5pt}
\pgfplotstableread[col sep=comma]{figures/impl_speed_seconds.csv}\impldata
\pgfplotstabletypeset[
  columns={dataset,batched_hartigan,hartigan_cpp,lederman_hartigan},
  columns/dataset/.style={
    column name={Dataset},
    column type={l},
    string type,
  },
  columns/batched_hartigan/.style={
    column name={Batched},
    string type,
  },
  columns/hartigan_cpp/.style={
    column name={C++},
    string type,
  },
  columns/lederman_hartigan/.style={
    column name={Lederman},
    string type,
  },
  every head row/.style={
    before row={\toprule
      & \multicolumn{3}{c}{Wall time (seconds)} \\
      \cmidrule(lr){2-4}},
    after row=\midrule,
  },
  every last row/.style={after row=\bottomrule},
]{\impldata}
\end{wraptable}

We compare our Batched Hartigan against two alternative
 Hartigan implementations:
the sequential Hartigan from~\cite{Slonim:2013:HKV:2540128.2540369}
(C++ with Cython bindings; \texttt{hartigan\_cpp}) and the
implementation of~\citet{lederman2026}
(Python with Numba JIT; \texttt{lederman}).
All three produce Hartigan fixed points; the difference is purely
in per-epoch throughput.

\begin{figure}[!h]
\centering
\begin{tikzpicture}
\begin{axis}[
  results,
  width=\columnwidth,
  height=0.3\columnwidth,
  xmode=log, ymode=log,
  xlabel={Work per epoch ($N \times D$)},
  ylabel={Wall time (s)},
  legend style={
    at={(0.03,0.97)}, anchor=north west,
    font=\footnotesize,
  },
  grid=both,
  grid style={dashed, gray!30},
  mark size=2.5pt,
]

\addplot[
  only marks,
  mark=*,
  color=Set1-A,
] table [
  col sep=comma,
  x=ND, y=batched_hartigan,
]{figures/impl_speed_agg_clean.csv};
\addlegendentry{Batched Hartigan (ours)}

\addplot[
  only marks,
  mark=square*,
  color=Set1-B,
] table [
  col sep=comma,
  x=ND, y=hartigan_cpp,
]{figures/impl_speed_agg_clean.csv};
\addlegendentry{C++ Hartigan}

\addplot[
  only marks,
  mark=triangle*,
  color=Set1-C,
] table [
  col sep=comma,
  x=ND, y=lederman_hartigan,
]{figures/impl_speed_agg_clean.csv};
\addlegendentry{Lederman Hartigan}

\end{axis}
\end{tikzpicture}
\caption{Wall-clock time vs.\ per-epoch work volume $N \times D$
  (median across $k$, single-threaded).  Batched Hartigan's GEMM
  acceleration and triangle-inequality pruning yield orders-of-magnitude
  speedups at large scale.}\label{fig:impl-speed}
\end{figure}
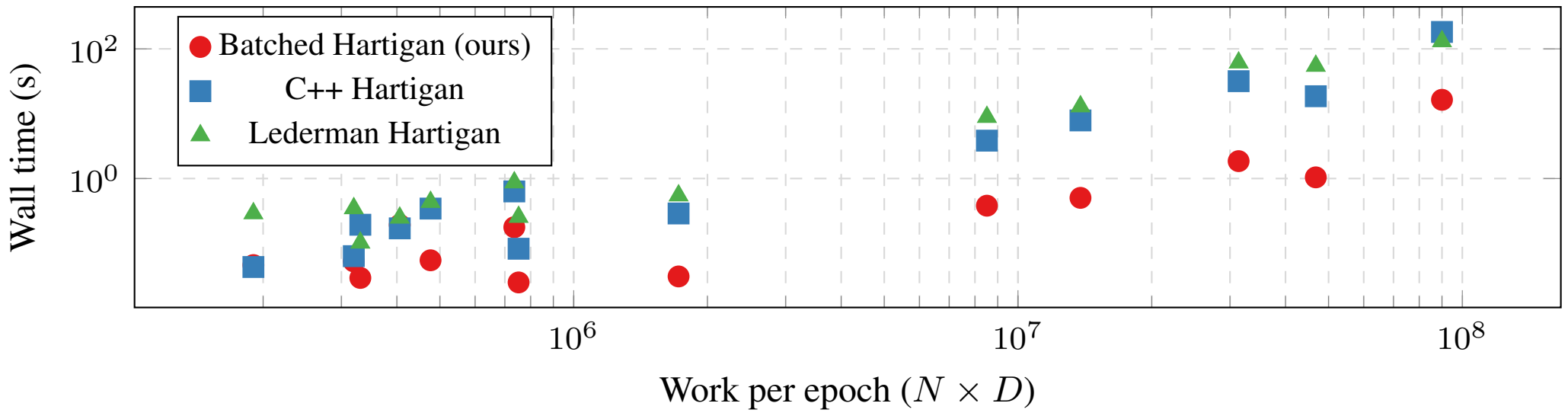

Table~\ref{tab:impl-speedup} gives per-dataset wall times.
Times are displayed directly in seconds.

Figure~\ref{fig:impl-speed} plots wall-clock time against
the per-epoch work volume $N \times D$ (aggregated as the median
across $k$ values per dataset).
Batched Hartigan is fastest on every dataset except the two
smallest, where GEMM overhead dominates.  The gap widens with
scale: on \textsc{covtype} ($N\!=\!581$k, $D\!=\!54$), Batched
Hartigan is $17\times$ faster than \texttt{hartigan\_cpp} and
$33\times$ faster than \texttt{lederman}.

%% file: sections/results_runtime.tex
\subsection{Runtime Comparison}\label{sec:runtime}

\begin{wraptable}[15]{r}{0.62\textwidth}
\vspace{-24pt}
\centering
\caption{Median wall time across $k$ values, in seconds.
  }%
\label{tab:runtime}
\small
\setlength{\tabcolsep}{5pt}
\pgfplotstableread[col sep=comma]{figures/runtime_seconds.csv}\runtimedata
\pgfplotstabletypeset[
  columns={dataset,lloyd,elkan,hamerly,batched_hartigan},
  columns/dataset/.style={
    column name={Dataset},
    column type={l},
    string type,
  },
  columns/lloyd/.style={
    column name={Lloyd},
    string type,
    assign cell content/.code={%
      \def\temp{##1}%
      \ifx\temp\empty
        \pgfkeyssetvalue{/pgfplots/table/@cell content}{---}%
      \else
        \pgfkeyssetvalue{/pgfplots/table/@cell content}{##1}%
      \fi
    },
  },
  columns/elkan/.style={
    column name={Elkan},
    string type,
  },
  columns/hamerly/.style={
    column name={Hamerly},
    string type,
  },
  columns/batched_hartigan/.style={
    column name={Batched H.},
    string type,
  },
  every head row/.style={
    before row={\toprule
      & \multicolumn{4}{c}{Wall time (seconds)} \\
      \cmidrule(lr){2-5}},
    after row=\midrule,
  },
  every last row/.style={after row=\bottomrule},
]{\runtimedata}
\end{wraptable}

We now compare Batched Hartigan to the accelerated Lloyd-family
algorithms: Elkan~\citep{elkan2003} (scikit-learn's default),
Hamerly~\citep{hamerly2010} (C++ implementation
of~\citet{drake2012}), and standard Lloyd~\citep{lloyd1982}
(our NumPy implementation).  All run single-threaded.

Table~\ref{tab:runtime} reports per-dataset median wall times.
The corresponding per-$k$ results are reported in
\autoref{app:per-k-results}.
At high $D$ (e.g.\ \textsc{mnist}, $D\!=\!780$; \textsc{aloi},
$D\!=\!128$) Batched Hartigan is competitive with or faster than
Elkan despite producing strictly better fixed points.
At low~$D$, Hamerly's cheaper $O(ND)$ bound update gives it an
edge.  Lloyd is consistently the slowest. This demonstates that our method is competitive with real-world $k$-means implementations in speed for the first time, enabling further study and experimentation.

%% file: sections/appendix_worked_example.tex
\section{Worked Example}\label{app:worked-example}

This section follows one batch from screening through the exact fallback.
The example uses $N=54$ two-dimensional observations divided among $k=3$
clusters. At the start of the batch, the cluster sizes and centroids are
\[
  (n_1,n_2,n_3)=(20,4,30),
  \qquad
  \boldsymbol{\mu}_1=(2,1),\quad
  \boldsymbol{\mu}_2=(5,4),\quad
  \boldsymbol{\mu}_3=(8,2).
\]
The symbols $\mathbf{x}_1,\ldots,\mathbf{x}_8$ denote the eight candidate
observations considered for this batch of size $B=8$; they do not denote
all $N$ observations. The other 46 observations determine the cluster
sizes and centroids above but are omitted from the plots.

\subsection{Step 1: Screen Stable Candidates}

The screening step decides whether an observation needs any new distance
calculation. Three candidates are already stable:
\[
\begin{array}{c|c|c|c}
\text{candidate} & \text{coordinates} & \text{current cluster}
  & (u_i,\ell_i)\\
\hline
\mathbf{x}_1 & (1.70,1.10) & C_1 & (0.4,3.5)\\
\mathbf{x}_2 & (5.20,4.10) & C_2 & (0.3,3.1)\\
\mathbf{x}_3 & (8.10,1.80) & C_3 & (0.3,3.0)
\end{array}
\]
Each satisfies $u_i<\ell_i$, so the assigned centroid remains closer than
any competing centroid under the stored cross-batch bounds. These three
observations are skipped. The five remaining candidates
$\mathbf{x}_4,\ldots,\mathbf{x}_8$ enter the batch.

\subsection{Step 2: Compute the Batch Distances}

The active observations are stacked as the five rows of
$X_{\mathcal B}$. One GEMM evaluates all $5\times3=15$ squared distances:
\[
  \widehat D
  =
  \|X_{\mathcal B}\|^2
  -2X_{\mathcal B}\boldsymbol{\mu}^{\mathsf T}
  +\|\boldsymbol{\mu}\|^2.
\]
\autoref{tab:worked-distances} gives both the coordinates and the resulting
rows of $\widehat D$. The ``current cluster'' column records the assignment
before this batch; it is not inferred from marker color or shape.

\begin{table}[!h]
\centering
\caption{The five active observations and the squared distances computed
by one GEMM.}
\label{tab:worked-distances}
\small
\begin{tabular}{cccrrr}
\toprule
Point & Coordinates & Current cluster
  & $\hat d_{i1}$ & $\hat d_{i2}$ & $\hat d_{i3}$ \\
\midrule
$\mathbf{x}_4$ & $(2.40,0.80)$ & $C_1$ & 0.20 & 17.00 & 32.80 \\
$\mathbf{x}_5$ & $(4.80,4.40)$ & $C_2$ & 19.40 & 0.20 & 16.00 \\
$\mathbf{x}_6$ & $(7.60,2.30)$ & $C_3$ & 33.05 & 9.65 & 0.25 \\
$\mathbf{x}_7$ & $(4.00,4.00)$ & $C_1$ & 13.00 & 1.00 & 20.00 \\
$\mathbf{x}_8$ & $(3.25,2.32)$ & $C_1$ & 3.30 & 5.88 & 22.66 \\
\bottomrule
\end{tabular}
\end{table}

\subsection{Step 3: Certify Safe Decisions}

The S1/S2 tests combine the precomputed distances with the current cluster
sizes and drift bounds. They certify four of the five active observations:
\begin{itemize}
  \item $\mathbf{x}_4$, $\mathbf{x}_5$, and $\mathbf{x}_6$ satisfy the
        safe-stay condition (S2): for each point, every candidate cluster's
        lower-bound Hartigan cost is at least the current cluster's
        upper-bound cost. Therefore they remain in $C_1$, $C_2$, and $C_3$,
        respectively.
  \item $\mathbf{x}_7$ satisfies the safe-move condition (S1): the
        upper-bound cost of moving to $C_2$ is below both the lower-bound
        cost of remaining in $C_1$ and the lower-bound cost of moving to
        any other candidate cluster. Therefore it moves from $C_1$ to
        $C_2$.
  \item $\mathbf{x}_8$ is not certified by either test and is deferred to
        the sequential fallback.
\end{itemize}

The ambiguity of $\mathbf{x}_8$ can be seen numerically. The move of
$\mathbf{x}_7$ permits centroid drifts
$\delta_1=0.190$ and $\delta_2=0.200$. Applying the Hartigan weights and
the distance bounds from \autoref{lem:dist-bounds} gives
\[
  [H_1^{\mathrm{lo}},H_1^{\mathrm{hi}}]=[2.80,4.25],
  \qquad
  [H_2^{\mathrm{lo}},H_2^{\mathrm{hi}}]=[4.13,5.75].
\]
The intervals overlap on $[4.13,4.25]$. Therefore the upper bound for
staying is not below the lower bound for moving, so S2 cannot certify a
stay; S1 also cannot certify a move.

\begin{figure}[!h]
\centering

\begin{tikzpicture}
\begin{axis}[
  width=0.92\textwidth,
  height=4.0cm,
  xmin=0, xmax=10,
  ymin=0, ymax=5.4,
  xtick={0,2,4,6,8,10},
  ytick={0,1,2,3,4,5},
  xlabel={Feature 1},
  ylabel={Feature 2},
  xlabel style={
    at={(axis description cs:0.5,-0.12)},
    anchor=north
  },
  title={\textbf{(a) Screening: eight candidates become five active points}},
  title style={font=\small},
  grid=major,
  grid style={gray!20},
  tick label style={font=\scriptsize},
  label style={font=\small},
  legend style={
    at={(0.5,-0.40)},
    anchor=north,
    legend columns=3,
    font=\scriptsize,
    draw=none
  },
  clip=false
]
\addplot[
  only marks,
  mark=x,
  mark size=4pt,
  very thick,
  gray
] coordinates {
  (1.7,1.1) (5.2,4.1) (8.1,1.8)
};
\addlegendentry{screened candidate}
\addplot[
  only marks,
  mark=o,
  mark size=3.5pt,
  very thick,
  MidnightBlue
] coordinates {
  (2.4,0.8) (4.8,4.4) (7.6,2.3) (4.0,4.0) (3.25,2.32)
};
\addlegendentry{sent to GEMM}
\addplot[
  only marks,
  mark=triangle*,
  mark size=5pt,
  black
] coordinates {
  (2,1) (5,4) (8,2)
};
\addlegendentry{centroid}

\node[font=\scriptsize, text=gray, anchor=south east]
  at (axis cs:1.7,1.1) {$\mathbf{x}_1$};
\node[font=\scriptsize, text=gray, anchor=south west]
  at (axis cs:5.2,4.1) {$\mathbf{x}_2$};
\node[font=\scriptsize, text=gray, anchor=north west]
  at (axis cs:8.1,1.8) {$\mathbf{x}_3$};
\node[font=\scriptsize, text=MidnightBlue, anchor=north west]
  at (axis cs:2.4,0.8) {$\mathbf{x}_4$};
\node[font=\scriptsize, text=MidnightBlue, anchor=south east]
  at (axis cs:4.8,4.4) {$\mathbf{x}_5$};
\node[font=\scriptsize, text=MidnightBlue, anchor=south east]
  at (axis cs:7.6,2.3) {$\mathbf{x}_6$};
\node[font=\scriptsize, text=MidnightBlue, anchor=south east]
  at (axis cs:4.0,4.0) {$\mathbf{x}_7$};
\node[font=\scriptsize, text=MidnightBlue, anchor=south west]
  at (axis cs:3.25,2.32) {$\mathbf{x}_8$};
\node[font=\scriptsize, anchor=north east]
  at (axis cs:2,1) {$\boldsymbol{\mu}_1$};
\node[font=\scriptsize, anchor=north west]
  at (axis cs:5,4) {$\boldsymbol{\mu}_2$};
\node[font=\scriptsize, anchor=south west]
  at (axis cs:8,2) {$\boldsymbol{\mu}_3$};
\end{axis}
\end{tikzpicture}

\vspace{4ex}

\begin{tikzpicture}
\begin{axis}[
  width=0.92\textwidth,
  height=4.0cm,
  xmin=0, xmax=10,
  ymin=0, ymax=5.4,
  xtick={0,2,4,6,8,10},
  ytick={0,1,2,3,4,5},
  xlabel={Feature 1},
  ylabel={Feature 2},
  xlabel style={
    at={(axis description cs:0.5,-0.12)},
    anchor=north
  },
  title={\textbf{(b) Certification: four safe decisions and one fallback}},
  title style={font=\small},
  grid=major,
  grid style={gray!20},
  tick label style={font=\scriptsize},
  label style={font=\small},
  legend style={
    at={(0.5,-0.40)},
    anchor=north,
    legend columns=4,
    font=\scriptsize,
    draw=none
  },
  clip=false
]
\addplot[
  only marks,
  mark=*,
  mark size=3.5pt,
  ForestGreen
] coordinates {
  (2.4,0.8) (4.8,4.4) (7.6,2.3)
};
\addlegendentry{S2: safe stay}
\addplot[
  only marks,
  mark=star,
  mark size=6pt,
  BrickRed
] coordinates {
  (4.0,4.0)
};
\addlegendentry{S1: safe move}
\addplot[
  only marks,
  mark=diamond*,
  mark size=4pt,
  Orange
] coordinates {
  (3.25,2.32)
};
\addlegendentry{ambiguous}
\addplot[
  only marks,
  mark=triangle*,
  mark size=5pt,
  black
] coordinates {
  (2,1) (5,4) (8,2)
};
\addlegendentry{centroid}

\node[font=\scriptsize, text=ForestGreen, anchor=north west]
  at (axis cs:2.4,0.8) {$\mathbf{x}_4$};
\node[font=\scriptsize, text=ForestGreen, anchor=south east]
  at (axis cs:4.8,4.4) {$\mathbf{x}_5$};
\node[font=\scriptsize, text=ForestGreen, anchor=south east]
  at (axis cs:7.6,2.3) {$\mathbf{x}_6$};
\node[font=\scriptsize, text=BrickRed, anchor=south east]
  at (axis cs:4.0,4.0) {$\mathbf{x}_7$};
\node[font=\scriptsize, text=Orange, anchor=south west]
  at (axis cs:3.25,2.32) {$\mathbf{x}_8$};
\node[font=\scriptsize, anchor=north east]
  at (axis cs:2,1) {$\boldsymbol{\mu}_1$};
\node[font=\scriptsize, anchor=north west]
  at (axis cs:5,4) {$\boldsymbol{\mu}_2$};
\node[font=\scriptsize, anchor=south west]
  at (axis cs:8,2) {$\boldsymbol{\mu}_3$};
\draw[-{Stealth[length=2mm]}, very thick, BrickRed]
  (axis cs:4.0,4.0)
  -- node[above, font=\scriptsize, text=BrickRed] {$C_1\to C_2$}
  (axis cs:5,4);
\end{axis}
\end{tikzpicture}

\caption{The eight candidate observations in the worked example.
\textbf{(a)} Gray crosses are removed by cross-batch screening, while blue
open circles proceed to the GEMM. \textbf{(b)} Green circles are certified
S2 stays, the red star is the S1 move from $C_1$ to $C_2$, and the orange
diamond requires exact fallback. Black triangles are centroids in both
panels. The axes give the two feature coordinates.}
\label{fig:worked-example}
\end{figure}
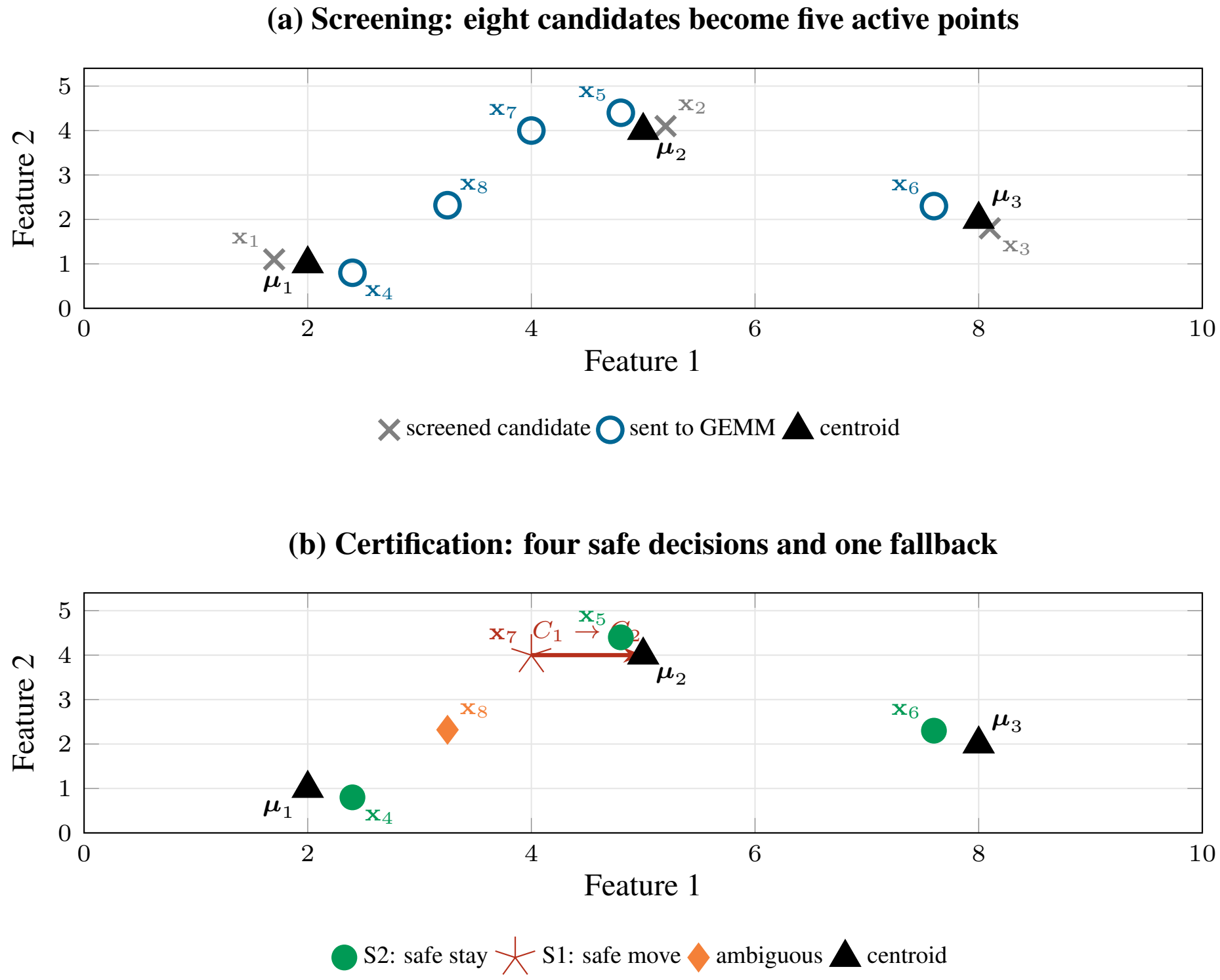

\autoref{fig:worked-example} visualizes Steps 1 and 3. In both panels,
horizontal and vertical position are simply the observation's first and
second feature values. Color and shape encode only the computational
status stated in each legend, not cluster membership.

\subsection{Step 4: Apply the Certified Move}

The three S2 decisions leave their assignments unchanged. Applying the
single S1 decision moves $\mathbf{x}_7=(4,4)$ from $C_1$ to $C_2$, so the
sizes change from $(20,4,30)$ to $(19,5,30)$. Both affected centroids must
be updated:
\[
\begin{aligned}
\boldsymbol{\mu}_1'
  &= \frac{20(2,1)-(4,4)}{19}
   = \left(\frac{36}{19},\frac{16}{19}\right)
   \approx (1.895,0.842),
&
\|\boldsymbol{\mu}_1'-\boldsymbol{\mu}_1\|
  &\approx 0.190,\\
\boldsymbol{\mu}_2'
  &= \frac{4(5,4)+(4,4)}{5}
   = (4.8,4),
&
\|\boldsymbol{\mu}_2'-\boldsymbol{\mu}_2\|
  &= 0.200.
\end{aligned}
\]
This is the bulk-update result: the three certified stays require no
change, and the one certified move updates only its source and destination
clusters.

\subsection{Step 5: Resolve the Ambiguous Point Exactly}

The fallback now processes $\mathbf{x}_8=(3.25,2.32)$ against the updated
centroids and sizes. The exact squared distances are
\[
\begin{aligned}
\|\mathbf{x}_8-\boldsymbol{\mu}_1'\|^2 &= 4.021,\\
\|\mathbf{x}_8-\boldsymbol{\mu}_2'\|^2 &= 5.225,\\
\|\mathbf{x}_8-\boldsymbol{\mu}_3\|^2  &= 22.665.
\end{aligned}
\]
Because $\mathbf{x}_8$ currently belongs to $C_1$, staying uses the
removal weight $19/(19-1)$. Moving to $C_2$ or $C_3$ uses the corresponding
insertion weight. The exact Hartigan costs are therefore
\[
\begin{aligned}
\Delta_H^2(\mathbf{x}_8,C_1)
  &= \frac{19}{18}(4.021) = 4.244,\\
\Delta_H^2(\mathbf{x}_8,C_2)
  &= \frac{5}{6}(5.225) = 4.354,\\
\Delta_H^2(\mathbf{x}_8,C_3)
  &= \frac{30}{31}(22.665) = 21.934.
\end{aligned}
\]
The current-cluster cost $4.244$ is the smallest, so the exact fallback
keeps $\mathbf{x}_8$ in $C_1$. The batch is complete: three observations
were screened before GEMM, four decisions were certified from the
precomputed distances and bounds, and only one observation required exact
sequential work.

%% file: sections/appendix_bounds_figure.tex
\section{Visual Depiction of Batched Hartigan Bounds}\label{app:bounds-figure}

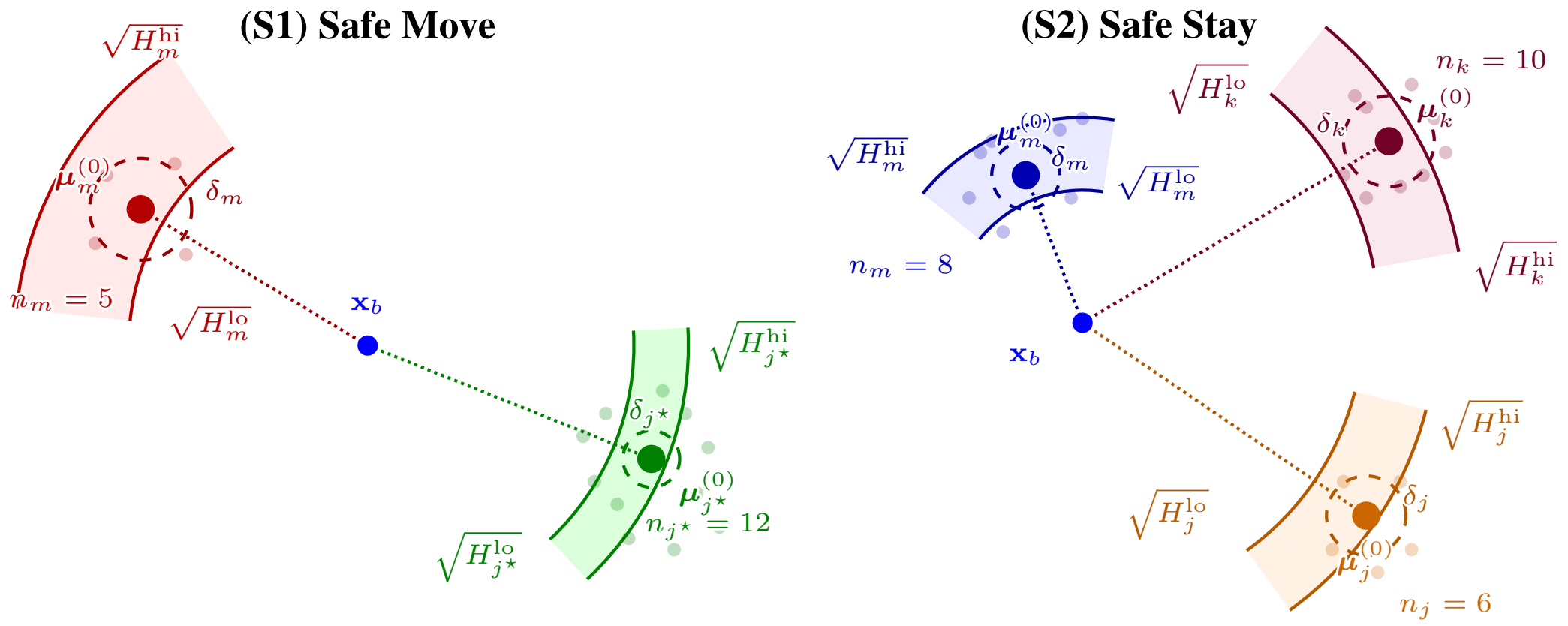
\begin{figure}[!h]
\centering
\begin{tikzpicture}[
  >=Stealth,
  point/.style={circle, fill, inner sep=1.8pt},
  centroid/.style={circle, fill, inner sep=2.5pt},
  ghost/.style={circle, fill, opacity=0.25, inner sep=1.2pt},
  drift circle/.style={dashed, thick},
  dist line/.style={thick, densely dotted},
  eqlabel/.style={font=\scriptsize, inner sep=0pt, outer sep=1pt},
]

\begin{scope}[shift={(0,0)}]

  \node[font=\bfseries] at (3, 5.8) {(S1) Safe Move};

  \coordinate (xb) at (3, 3);        %
  \coordinate (mu_m) at (1.0, 4.2);  %
  \coordinate (mu_j) at (5.5, 2.0);  %

  \def\driftm{0.45}
  \def\driftj{0.25}

  \pgfmathsetmacro{\distm}{veclen(3-1.0, 3-4.2)}
  \pgfmathsetmacro{\distj}{veclen(3-5.5, 3-2.0)}

  \def\wm{1.25}    %
  \def\wj{0.923}   %

  \pgfmathsetmacro{\distmLo}{\distm - \driftm}
  \pgfmathsetmacro{\distmHi}{\distm + \driftm}
  \pgfmathsetmacro{\distjLo}{\distj - \driftj}
  \pgfmathsetmacro{\distjHi}{\distj + \driftj}

  \pgfmathsetmacro{\HmLo}{sqrt(\wm)*\distmLo}
  \pgfmathsetmacro{\HmHi}{sqrt(\wm)*\distmHi}
  \pgfmathsetmacro{\HjLo}{sqrt(\wj)*\distjLo}
  \pgfmathsetmacro{\HjHi}{sqrt(\wj)*\distjHi}

  \pgfmathsetmacro{\angm}{atan2(4.2-3, 1.0-3)}
  \fill[red!12, opacity=0.7]
    (xb) ++(\angm-25:\HmLo) arc (\angm-25:\angm+25:\HmLo)
    -- ++(\angm+25:\HmHi-\HmLo) arc (\angm+25:\angm-25:\HmHi)
    -- cycle;
  \draw[red!70!black, thick] (xb) ++(\angm-25:\HmLo) arc (\angm-25:\angm+25:\HmLo);
  \draw[red!70!black, thick] (xb) ++(\angm-25:\HmHi) arc (\angm-25:\angm+25:\HmHi);

  \pgfmathsetmacro{\angj}{atan2(2.0-3, 5.5-3)}
  \fill[green!20, opacity=0.7]
    (xb) ++(\angj-25:\HjLo) arc (\angj-25:\angj+25:\HjLo)
    -- ++(\angj+25:\HjHi-\HjLo) arc (\angj+25:\angj-25:\HjHi)
    -- cycle;
  \draw[green!50!black, thick] (xb) ++(\angj-25:\HjLo) arc (\angj-25:\angj+25:\HjLo);
  \draw[green!50!black, thick] (xb) ++(\angj-25:\HjHi) arc (\angj-25:\angj+25:\HjHi);

  \draw[drift circle, red!60!black] (mu_m) circle (\driftm);
  \draw[drift circle, green!50!black] (mu_j) circle (\driftj);

  \node[ghost, red!70!black] at (0.7, 4.5) {};
  \node[ghost, red!70!black] at (1.3, 4.6) {};
  \node[ghost, red!70!black] at (0.6, 3.9) {};
  \node[ghost, red!70!black] at (1.4, 3.8) {};

  \node[ghost, green!50!black] at (5.2, 1.6) {};
  \node[ghost, green!50!black] at (5.8, 2.4) {};
  \node[ghost, green!50!black] at (5.9, 1.7) {};
  \node[ghost, green!50!black] at (5.1, 2.4) {};
  \node[ghost, green!50!black] at (6.0, 2.1) {};
  \node[ghost, green!50!black] at (5.3, 1.3) {};
  \node[ghost, green!50!black] at (5.7, 1.2) {};
  \node[ghost, green!50!black] at (5.0, 1.8) {};
  \node[ghost, green!50!black] at (5.6, 2.6) {};
  \node[ghost, green!50!black] at (6.1, 1.4) {};
  \node[ghost, green!50!black] at (4.9, 2.2) {};

  \draw[dist line, red!60!black] (xb) -- (mu_m);
  \draw[dist line, green!50!black] (xb) -- (mu_j);

  \node[point, blue] at (xb) {};
  \node[centroid, red!70!black] at (mu_m) {};
  \node[centroid, green!50!black] at (mu_j) {};

  \node[eqlabel, text=blue] at ($(xb)+(0,0.35)$) {\contour{white}{$\mathbf{x}_b$}};
  \node[eqlabel, text=red!70!black] at ($(mu_m)+(-0.5,0.3)$)
    {\contour{white}{$\boldsymbol{\mu}_m^{(0)}$}};
  \node[eqlabel, text=green!50!black] at ($(mu_j)+(0.5,-0.3)$)
    {\contour{white}{$\boldsymbol{\mu}_{j^\star}^{(0)}$}};

  \node[eqlabel, text=red!60!black] at ($(mu_m)+(0.75,0.15)$)
    {\contour{white}{$\delta_m$}};
  \node[eqlabel, text=green!50!black] at ($(mu_j)+(0.0,0.4)$)
    {\contour{white}{$\delta_{j^\star}$}};

  \node[eqlabel, text=red!70!black, anchor=west] at
    (1.2, 3.2) {\contour{white}{$\sqrt{H_m^{\mathrm{lo}}}$}};
  \node[eqlabel, text=red!70!black, anchor=south] at
    (1.0, 5.5) {\contour{white}{$\sqrt{H_m^{\mathrm{hi}}}$}};
  \node[eqlabel, text=green!50!black, anchor=north east] at
    (4.4, 1.4) {\contour{white}{$\sqrt{H_{j^\star}^{\mathrm{lo}}}$}};
  \node[eqlabel, text=green!50!black, anchor=west] at
    (5.95, 3.0) {\contour{white}{$\sqrt{H_{j^\star}^{\mathrm{hi}}}$}};

  \node[eqlabel, text=red!70!black] at (0.3, 3.4) {\contour{white}{$n_m = 5$}};
  \node[eqlabel, text=green!50!black] at (6.0, 1.4) {\contour{white}{$n_{j^\star} = 12$}};

\end{scope}

\begin{scope}[shift={(6.8,0)}]

  \node[font=\bfseries] at (3, 5.8) {(S2) Safe Stay};

  \coordinate (xb2) at (2.5, 3.2);
  \coordinate (mu_m2) at (2.0, 4.5);
  \coordinate (mu_j2) at (5.0, 1.5);
  \coordinate (mu_k2) at (5.2, 4.8);

  \def\driftmm{0.3}
  \def\driftjj{0.35}
  \def\driftkk{0.4}

  \pgfmathsetmacro{\distmm}{veclen(2.5-2.0, 3.2-4.5)}
  \pgfmathsetmacro{\distjj}{veclen(2.5-5.0, 3.2-1.5)}
  \pgfmathsetmacro{\distkk}{veclen(2.5-5.2, 3.2-4.8)}

  \def\wmm{1.143}    %
  \def\wjj{0.857}    %
  \def\wkk{0.909}    %

  \pgfmathsetmacro{\distmmLo}{\distmm - \driftmm}
  \pgfmathsetmacro{\distmmHi}{\distmm + \driftmm}
  \pgfmathsetmacro{\distjjLo}{\distjj - \driftjj}
  \pgfmathsetmacro{\distjjHi}{\distjj + \driftjj}
  \pgfmathsetmacro{\distkkLo}{\distkk - \driftkk}
  \pgfmathsetmacro{\distkkHi}{\distkk + \driftkk}

  \pgfmathsetmacro{\HmmLo}{sqrt(\wmm)*\distmmLo}
  \pgfmathsetmacro{\HmmHi}{sqrt(\wmm)*\distmmHi}
  \pgfmathsetmacro{\HjjLo}{sqrt(\wjj)*\distjjLo}
  \pgfmathsetmacro{\HjjHi}{sqrt(\wjj)*\distjjHi}
  \pgfmathsetmacro{\HkkLo}{sqrt(\wkk)*\distkkLo}
  \pgfmathsetmacro{\HkkHi}{sqrt(\wkk)*\distkkHi}

  \pgfmathsetmacro{\angmm}{atan2(4.5-3.2, 2.0-2.5)}
  \fill[blue!12, opacity=0.7]
    (xb2) ++(\angmm-30:\HmmLo) arc (\angmm-30:\angmm+30:\HmmLo)
    -- ++(\angmm+30:\HmmHi-\HmmLo) arc (\angmm+30:\angmm-30:\HmmHi)
    -- cycle;
  \draw[blue!60!black, thick] (xb2) ++(\angmm-30:\HmmLo) arc (\angmm-30:\angmm+30:\HmmLo);
  \draw[blue!60!black, thick] (xb2) ++(\angmm-30:\HmmHi) arc (\angmm-30:\angmm+30:\HmmHi);

  \pgfmathsetmacro{\angjj}{atan2(1.5-3.2, 5.0-2.5)}
  \fill[orange!15, opacity=0.7]
    (xb2) ++(\angjj-20:\HjjLo) arc (\angjj-20:\angjj+20:\HjjLo)
    -- ++(\angjj+20:\HjjHi-\HjjLo) arc (\angjj+20:\angjj-20:\HjjHi)
    -- cycle;
  \draw[orange!70!black, thick] (xb2) ++(\angjj-20:\HjjLo) arc (\angjj-20:\angjj+20:\HjjLo);
  \draw[orange!70!black, thick] (xb2) ++(\angjj-20:\HjjHi) arc (\angjj-20:\angjj+20:\HjjHi);

  \pgfmathsetmacro{\angkk}{atan2(4.8-3.2, 5.2-2.5)}
  \fill[purple!12, opacity=0.7]
    (xb2) ++(\angkk-20:\HkkLo) arc (\angkk-20:\angkk+20:\HkkLo)
    -- ++(\angkk+20:\HkkHi-\HkkLo) arc (\angkk+20:\angkk-20:\HkkHi)
    -- cycle;
  \draw[purple!60!black, thick] (xb2) ++(\angkk-20:\HkkLo) arc (\angkk-20:\angkk+20:\HkkLo);
  \draw[purple!60!black, thick] (xb2) ++(\angkk-20:\HkkHi) arc (\angkk-20:\angkk+20:\HkkHi);

  \draw[drift circle, blue!60!black] (mu_m2) circle (\driftmm);
  \draw[drift circle, orange!70!black] (mu_j2) circle (\driftjj);
  \draw[drift circle, purple!60!black] (mu_k2) circle (\driftkk);

  \node[ghost, blue!70!black] at (1.7, 4.8) {};
  \node[ghost, blue!70!black] at (2.3, 4.9) {};
  \node[ghost, blue!70!black] at (1.5, 4.3) {};
  \node[ghost, blue!70!black] at (2.4, 4.3) {};
  \node[ghost, blue!70!black] at (1.8, 4.0) {};
  \node[ghost, blue!70!black] at (2.5, 5.0) {};
  \node[ghost, blue!70!black] at (1.6, 4.7) {};

  \node[ghost, orange!80!black] at (4.7, 1.2) {};
  \node[ghost, orange!80!black] at (5.3, 1.8) {};
  \node[ghost, orange!80!black] at (5.4, 1.2) {};
  \node[ghost, orange!80!black] at (4.8, 1.8) {};
  \node[ghost, orange!80!black] at (5.1, 1.0) {};

  \node[ghost, purple!70!black] at (5.0, 5.1) {};
  \node[ghost, purple!70!black] at (5.5, 4.5) {};
  \node[ghost, purple!70!black] at (5.6, 5.0) {};
  \node[ghost, purple!70!black] at (4.8, 4.5) {};
  \node[ghost, purple!70!black] at (5.4, 5.3) {};
  \node[ghost, purple!70!black] at (5.0, 4.3) {};
  \node[ghost, purple!70!black] at (5.7, 4.7) {};
  \node[ghost, purple!70!black] at (4.9, 5.2) {};
  \node[ghost, purple!70!black] at (5.3, 4.4) {};

  \draw[dist line, blue!60!black] (xb2) -- (mu_m2);
  \draw[dist line, orange!70!black] (xb2) -- (mu_j2);
  \draw[dist line, purple!60!black] (xb2) -- (mu_k2);

  \node[point, blue] at (xb2) {};
  \node[centroid, blue!70!black] at (mu_m2) {};
  \node[centroid, orange!80!black] at (mu_j2) {};
  \node[centroid, purple!60!black] at (mu_k2) {};

  \node[eqlabel, text=blue] at ($(xb2)+(-0.5,-0.3)$) {\contour{white}{$\mathbf{x}_b$}};
  \node[eqlabel, text=blue!70!black] at ($(mu_m2)+(0,0.4)$)
    {\contour{white}{$\boldsymbol{\mu}_m^{(0)}$}};
  \node[eqlabel, text=orange!80!black] at ($(mu_j2)+(0,-0.4)$)
    {\contour{white}{$\boldsymbol{\mu}_j^{(0)}$}};
  \node[eqlabel, text=purple!60!black] at ($(mu_k2)+(0.5,0.3)$)
    {\contour{white}{$\boldsymbol{\mu}_k^{(0)}$}};

  \node[eqlabel, text=blue!60!black] at ($(mu_m2)+(0.4,0.15)$)
    {\contour{white}{$\delta_m$}};
  \node[eqlabel, text=orange!70!black] at ($(mu_j2)+(0.45,0.15)$)
    {\contour{white}{$\delta_j$}};
  \node[eqlabel, text=purple!60!black] at ($(mu_k2)+(-0.5,0.15)$)
    {\contour{white}{$\delta_k$}};

  \node[eqlabel, text=blue!60!black, anchor=south east] at
    (1.0, 4.5) {\contour{white}{$\sqrt{H_m^{\mathrm{hi}}}$}};
  \node[eqlabel, text=blue!60!black, anchor=south west] at
    (2.75, 4.25) {\contour{white}{$\sqrt{H_m^{\mathrm{lo}}}$}};
  \node[eqlabel, text=orange!70!black, anchor=south east] at
    (3.65, 1.25) {\contour{white}{$\sqrt{H_j^{\mathrm{lo}}}$}};
  \node[eqlabel, text=orange!70!black, anchor=west] at
    (5.6, 2.3) {\contour{white}{$\sqrt{H_j^{\mathrm{hi}}}$}};
  \node[eqlabel, text=purple!60!black, anchor=south east] at
    (4.0, 5.0) {\contour{white}{$\sqrt{H_k^{\mathrm{lo}}}$}};
  \node[eqlabel, text=purple!60!black, anchor=west] at
    (5.9, 3.7) {\contour{white}{$\sqrt{H_k^{\mathrm{hi}}}$}};

  \node[eqlabel, text=blue!70!black] at (0.9, 3.7) {\contour{white}{$n_m = 8$}};
  \node[eqlabel, text=orange!80!black] at (5.7, 0.7) {\contour{white}{$n_j = 6$}};
  \node[eqlabel, text=purple!60!black] at (6.1, 5.5) {\contour{white}{$n_k = 10$}};

\end{scope}

\end{tikzpicture}
\caption{Geometric interpretation of Hartigan cost bounds.
  Dashed circles show centroid drift uncertainty $\delta_j$
  (the true centroid lies somewhere inside).
  Colored bands show the range of possible
  $\sqrt{H}$ values (Hartigan-weighted distance from
  $\mathbf{x}_b$); inner arcs are $\sqrt{H^{\mathrm{lo}}}$,
  outer arcs are $\sqrt{H^{\mathrm{hi}}}$.
  The Hartigan weight $\tfrac{n_j}{n_j \pm 1}$ shrinks bands
  for large clusters, making safe decisions more frequent.
  \textbf{Left:} $H_{j^\star}^{\mathrm{hi}} < H_m^{\mathrm{lo}}$
  so the point safely moves (worst-case candidate still better
  than best-case current).
  \textbf{Right:} $H_j^{\mathrm{lo}} \geq H_m^{\mathrm{hi}}$
  for all $j \neq m$, so the point safely stays (best-case
  candidate still worse than worst-case current).}%
\label{fig:bounds-diagram}
\end{figure}

Figure~\ref{fig:bounds-diagram} illustrates the geometric
interpretation of the Hartigan cost bounds $H^{\mathrm{lo}}$ and
$H^{\mathrm{hi}}$ used for safe reassignment decisions.
Each point $\mathbf{x}_b$ has a precomputed distance to each
centroid $\boldsymbol{\mu}_j^{(0)}$, but subsequent reassignments
of \emph{other} points cause centroids to drift by at most
$\delta_j$.  The triangle inequality then bounds the true
(unknown) distance within an interval, and the Hartigan weight
$\tfrac{n_j}{n_j \pm 1}$ converts these distance bounds into
cost bounds.  Geometrically, each cluster's possible Hartigan
cost occupies an annular band centered on $\mathbf{x}_b$: the
inner arc corresponds to $\sqrt{H^{\mathrm{lo}}}$ and the outer
arc to $\sqrt{H^{\mathrm{hi}}}$.

When the bands are well-separated, a decision can be made without
recomputing exact distances:
\textbf{(S1)}~if the worst-case candidate cost is still below
the best-case current cost, the point is safely moved;
\textbf{(S2)}~if the best-case cost of every candidate exceeds
the worst-case current cost, the point safely stays.
Large clusters produce narrower bands (the Hartigan weight
approaches~1), making safe decisions more likely.

%% file: sections/appendix_per_k_results.tex
\section{Per-k Runtime Results}\label{app:per-k-results}

This section reports the $k$-dependence underlying the aggregate runtime
comparison in \autoref{sec:runtime}.  We use the eight datasets with
$N \geq 50{,}000$ and complete Lloyd timings.

\begin{figure}[!h]
\centering
\begin{tikzpicture}
\begin{axis}[
  width=0.75\textwidth,
  height=0.52\textwidth,
  xlabel={Number of clusters ($k$)},
  ylabel={Dataset},
  xtick={0,1,2,3,4,5,6},
  xticklabels={2,4,8,16,32,64,128},
  ytick={0,1,2,3,4,5,6,7},
  yticklabels={Shuttle,cod-rna,mnist,connect-4,aloi,Sepsis,
    skin-nonskin,covtype},
  y dir=reverse,
  enlargelimits=false,
  axis on top,
  colormap/RdYlGn,
  point meta min=-1,
  point meta max=5,
  colorbar,
  colorbar style={
    ylabel={Speedup},
    ytick={-1,0,1,2,3,4,5},
    yticklabels={$0.5\times$,$1\times$,$2\times$,$4\times$,
      $8\times$,$16\times$,$32\times$},
  },
]
\addplot[
  matrix plot*,
  mesh/cols=7,
  point meta=explicit,
] table[
  col sep=comma,
  x=k_index,
  y=dataset_index,
  meta=log2_batched_hartigan,
] {figures/per_k_speedup.csv};
\end{axis}
\end{tikzpicture}
\caption{Batched Hartigan speedup relative to Lloyd by dataset and $k$.
Color encodes log-base-two speedup; $1\times$ is parity and higher is
faster.}
\label{fig:per-k-speedup}
\end{figure}
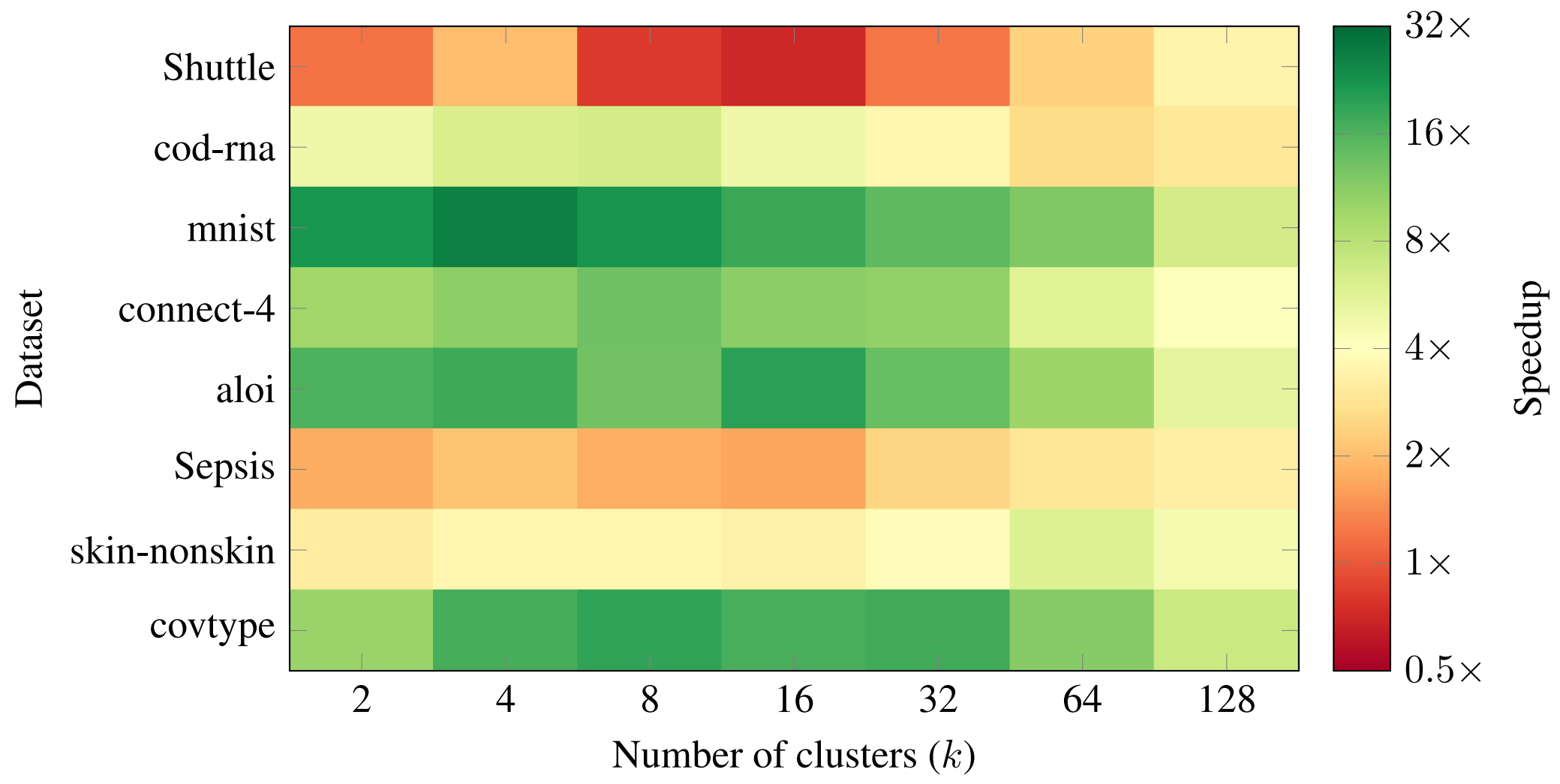

\autoref{fig:per-k-speedup} shows Batched Hartigan speedup relative to
Lloyd for each dataset and $k$.

\autoref{tab:per-k-speedup} summarizes all three comparisons over the
same datasets.
\begin{table}[!h]
\centering
\caption{Per-$k$ speedup relative to Lloyd.  Entries are mean
(minimum--maximum); higher is faster.}
\label{tab:per-k-speedup}
\small
\begin{tabular}{@{}lccc@{}}
\toprule
$k$ & Prior H.\ (C++) & Elkan & Batched H.\ (ours) \\
\midrule
2   & $2.6$ ($0.3$--$4.6$) & $12.8$ ($2.6$--$23.7$) & $8.5$ ($1.2$--$21.3$) \\
4   & $2.3$ ($0.3$--$5.9$) & $13.2$ ($4.0$--$23.4$) & $10.6$ ($2.0$--$26.0$) \\
8   & $1.6$ ($0.5$--$3.1$) & $12.0$ ($3.8$--$19.6$) & $9.9$ ($0.8$--$21.9$) \\
16  & $0.9$ ($0.2$--$1.2$) & $9.6$ ($2.8$--$16.8$) & $9.4$ ($0.7$--$19.4$) \\
32  & $0.7$ ($0.3$--$1.2$) & $8.6$ ($3.7$--$13.0$) & $8.4$ ($1.2$--$17.2$) \\
64  & $0.6$ ($0.2$--$0.9$) & $7.3$ ($4.5$--$11.4$) & $6.5$ ($2.3$--$11.8$) \\
128 & $0.4$ ($0.1$--$1.2$) & $6.0$ ($4.4$--$8.4$) & $4.6$ ($3.0$--$6.9$) \\
\bottomrule
\end{tabular}
\end{table}

Elkan and Batched Hartigan retain the same order of speedup at every
$k$.  Mean prior C++ Hartigan speedup falls below Lloyd from $k=16$
onward.

%% file: sections/appendix_datasets.tex
\section{Dataset Characteristics}\label{app:datasets}

This section reports the input scales and preprocessing used in
\autoref{sec:results}.  \autoref{tab:dataset-characteristics} lists all
datasets in increasing order of $N$.

\begin{table}[!h]
\centering
\caption{Dataset dimensions and total dense data elements.}
\label{tab:dataset-characteristics}
\small
\begin{tabular}{lrrr}
\toprule
Dataset & $N$ & $D$ & $N \times D$ \\
\midrule
phishing             & $11{,}055$  & 68  & 752K \\
MAGIC Gamma           & $19{,}020$  & 10  & 190K \\
Letter                & $20{,}000$  & 16  & 320K \\
Superconductivity     & $21{,}263$  & 81  & 1.7M \\
Shuttle               & $58{,}000$  & 7   & 406K \\
cod-rna               & $59{,}535$  & 8   & 476K \\
mnist                 & $60{,}000$  & 780 & 46.8M \\
connect-4             & $67{,}557$  & 126 & 8.5M \\
aloi                  & $108{,}000$ & 128 & 13.8M \\
Sepsis                & $110{,}341$ & 3   & 331K \\
skin-nonskin          & $245{,}057$ & 3   & 735K \\
covtype               & $581{,}012$ & 54  & 31.4M \\
SUSY                  & $5{,}000{,}000$ & 18 & 90M \\
\bottomrule
\end{tabular}
\end{table}

Features were used at source scale without additional normalization.
For UCI data, nonnumeric or entirely missing columns and rows containing
missing feature values were removed.  LIBSVM inputs were densified
without rescaling.

%% file: sections/appendix_sparse_results.tex
\section{Sparse Results}\label{app:sparse-results}

This section reports the density and runtime context for the sparse-data
quality results in \autoref{sec:efficacy}.  We evaluated 11 LIBSVM
datasets in CSR format using shared $k$-means++ initializations, three
trials, $k \in \{4,8,16,32,64,128,256\}$ when $k<N$, and at most 300
iterations.

\autoref{tab:sparse-results} reports dataset scale, the best observed
inertia ratio, and runtime at the corresponding $k$.
\begin{table}[!h]
\centering
\caption{Sparse benchmark results.  The inertia ratio is Batched
Hartigan divided by Elkan for the best observed trial.  Runtime speedup
is median Elkan time divided by median Batched Hartigan time over three
trials at the selected $k$; values above one favor Batched Hartigan.}
\label{tab:sparse-results}
\small
\begin{tabular}{@{}lrrrrrr@{}}
\toprule
Dataset & $N$ & $D$ & Density & $k$ & Inertia ratio & Runtime speedup \\
\midrule
breast-cancer & 683 & 10 & 1.0000 & 128 & 0.9504 & $1.51\times$ \\
leukemia & 38 & 7,129 & 1.0000 & 16 & 0.9134 & $0.25\times$ \\
duke & 44 & 7,129 & 1.0000 & 4 & 0.8201 & $0.29\times$ \\
bibtex & 7,395 & 1,836 & 0.0374 & 4 & 0.9639 & $1.73\times$ \\
sector & 6,412 & 55,197 & 0.0030 & 256 & 0.9866 & $1.57\times$ \\
news20 & 15,935 & 62,061 & 0.0013 & 8 & 0.9030 & $0.10\times$ \\
rcv1\_multi & 15,564 & 47,236 & 0.0014 & 256 & 0.9933 & $1.83\times$ \\
E2006-tfidf & 16,087 & 150,360 & 0.0083 & 32 & 0.9891 & $0.34\times$ \\
scotus & 7,800 & 126,405 & 0.0107 & 256 & 0.9775 & $0.23\times$ \\
Wiki10-31K & 20,762 & 104,374 & 0.0065 & 256 & 0.9845 & $0.28\times$ \\
real-sim & 72,309 & 20,958 & 0.0024 & 256 & 0.9794 & $1.25\times$ \\
\bottomrule
\end{tabular}
\end{table}

Across all 213 configurations, runtime speedup ranged from
$0.08\times$ to $3.20\times$, with median $0.54\times$; Batched
Hartigan was faster in 72 configurations.  Quality and runtime gains
therefore occur in different regimes.

%% file: sections/appendix_refinement.tex
\section{Lloyd Refinement}\label{app:refinement}

This section evaluates Batched Hartigan initialized from a converged
Lloyd solution.  This refinement is valid because Hartigan fixed points
form a subset of Lloyd fixed points~\citep{telgarsky2010}.

For each dataset and selected $k$, we ran five shared $k$-means++
initializations.  Elkan was run to a Lloyd fixed point; Batched Hartigan
was run both from the original initialization and from Elkan's converged
centers.  \autoref{tab:refinement} reports quality and refinement cost.
\begin{table}[!h]
\centering
\caption{Batched Hartigan as a Lloyd refinement.  Direct is the best
observed Batched-Hartigan-to-Elkan inertia ratio from the original
three-trial sweep; Refine is the median ratio over five refinement
trials.  Overhead is the median of
$100\,t_{\mathrm{refine}}/t_{\mathrm{Elkan}}$.}
\label{tab:refinement}
\footnotesize
\setlength{\tabcolsep}{3pt}
\begin{tabular}{@{}lrrrrrrr@{}}
\toprule
Dataset & $N$ & $D$ & Density & $k$ & Direct & Refine & Overhead \\
\midrule
breast-cancer & 683 & 10 & 1.0000 & 128 & 0.9504 & 0.9754 & 44.1\% \\
leukemia & 38 & 7,129 & 1.0000 & 16 & 0.9134 & 0.9407 & 305.4\% \\
duke & 44 & 7,129 & 1.0000 & 4 & 0.8201 & 0.9551 & 298.5\% \\
bibtex & 7,395 & 1,836 & 0.0374 & 4 & 0.9639 & 0.9864 & 75.7\% \\
sector & 6,412 & 55,197 & 0.0030 & 256 & 0.9866 & 0.9869 & 77.2\% \\
news20 & 15,935 & 62,061 & 0.0013 & 8 & 0.9030 & 0.9305 & 493.8\% \\
rcv1\_multi & 15,564 & 47,236 & 0.0014 & 256 & 0.9933 & 0.9934 & 63.5\% \\
E2006-tfidf & 16,087 & 150,360 & 0.0083 & 32 & 0.9891 & 1.0073 & 130.1\% \\
scotus & 7,800 & 126,405 & 0.0107 & 256 & 0.9775 & 0.9801 & 392.9\% \\
Wiki10-31K & 20,762 & 104,374 & 0.0065 & 256 & 0.9845 & 0.9917 & 362.7\% \\
real-sim & 72,309 & 20,958 & 0.0024 & 256 & 0.9794 & 0.9843 & 81.6\% \\
\bottomrule
\end{tabular}
\end{table}

Refinement lowered median inertia on 10 of 11 datasets, but its ratio
was higher than the selected direct-start result in every row.
Refinement added 44.1--493.8\% of one Elkan run.  We did not evaluate
matched-budget independent Lloyd restarts, so these results do not
establish a comparison with that strategy.

%% file: sections/appendix_ablation.tex
\section{Ablation Study}\label{app:ablation}

We conduct two ablation studies to validate the design of Batched Hartigan: (1)~a component-removal study showing that each algorithmic component contributes meaningfully, and (2)~a hyperparameter sensitivity study confirming that defaults are robust across diverse datasets. All experiments use single-threaded NumPy (no multi-core BLAS), tolerance $\tau=0$, maximum 1000 iterations, and are averaged over 2 trials with shared $k$-means++ initialization. Seven datasets are tested spanning $N{=}11\text{k}$ to $N{=}5\text{M}$.

\subsection{Component Removal}\label{app:ablation-components}

We test five ablated variants against the full Batched Hartigan algorithm:

\begin{itemize}
\item \textbf{No BuildBatch}: Screening is disabled; all $N$ points are treated as unsettled every epoch, eliminating the $O(N)$ savings from skipping converged points.
\item \textbf{No Parallel}: The bulk GEMM-based reassignment path is disabled; all unsettled points are processed by the sequential kernel with per-point $O(k)$ cost.
\item \textbf{No Sequential}: The drift-aware sequential bounds (S1/S2) are disabled; ambiguous points are resolved by brute-force exact-distance Hartigan reassignment.
\item \textbf{No Dynamic $B_{\max}$}: Batch size is fixed at $B_{\max}=4096$ (no EMA-based adaptation).
\item \textbf{No Dynamic $B_{\min}$}: Batch size is fixed at $B_{\min}=64$.
\end{itemize}

Table~\ref{tab:ablation-components} reports the wall-time ratio of each variant relative to the full algorithm (values ${>}1$ indicate slowdown). All variants converge to equivalent inertia (${<}0.01\%$ difference), confirming that the components affect \emph{speed}, not solution quality.

\begin{table}[!h]
\centering
\caption{Component ablation: wall-time ratio relative to full Batched Hartigan. Values ${>}1$ indicate slowdown. Bold marks the most impactful removal per row.}
\label{tab:ablation-components}
\small
\begin{tabular}{llrrrrr}
\toprule
Dataset & $k$ & No Build & No Par. & No Seq. & No Dyn.$_{\max}$ & No Dyn.$_{\min}$ \\
\midrule
phishing       &   8 & 1.48 & 0.71 & 0.97 & 0.95 & \textbf{5.10} \\
               &  32 & 1.26 & 0.80 & 1.05 & 1.00 & \textbf{3.63} \\
\midrule
Letter         &   8 & 1.86 & 0.69 & 0.95 & 0.98 & \textbf{6.26} \\
               &  32 & 1.58 & 0.80 & 0.87 & 0.89 & \textbf{5.94} \\
               & 128 & 2.00 & 0.92 & 1.00 & 1.03 & \textbf{3.47} \\
\midrule
mnist          &   8 & 1.55 & 0.93 & 1.32 & 1.00 & \textbf{2.73} \\
               &  32 & 1.11 & 0.93 & \textbf{1.51} & 1.00 & 1.90 \\
               & 128 & 1.34 & 1.05 & \textbf{1.53} & 1.01 & 1.42 \\
\midrule
connect-4      &   8 & 1.62 & 0.72 & 1.02 & 0.95 & \textbf{4.85} \\
               &  32 & 1.46 & 0.96 & 1.15 & 1.00 & \textbf{3.50} \\
               & 128 & 1.43 & 0.84 & 1.21 & 1.01 & \textbf{1.97} \\
\midrule
aloi           &   8 & 2.23 & 0.69 & 0.99 & 0.96 & \textbf{5.59} \\
               &  32 & 2.10 & 0.87 & 1.05 & 1.00 & \textbf{4.49} \\
               & 128 & 1.20 & 0.68 & 1.07 & 1.00 & \textbf{2.29} \\
\midrule
covtype        &   8 & 4.03 & 0.69 & 1.00 & 1.00 & \textbf{5.81} \\
               &  32 & \textbf{18.91} & 1.23 & 0.83 & 0.83 & 5.68 \\
               & 128 & \textbf{10.90} & 0.67 & 0.76 & 1.02 & 3.11 \\
\midrule
SUSY           &   8 & 3.12 & 0.87 & 1.02 & 1.00 & \textbf{8.13} \\
               &  32 & \textbf{9.55} & \textbf{1.27} & 0.99 & 1.00 & 8.93 \\
               & 128 & \textbf{4.73} & 0.94 & 0.98 & 0.99 & 3.82 \\
\bottomrule
\end{tabular}
\end{table}

\paragraph{Screening (BuildBatch) is consistently important.}
Removing screening forces reprocessing of all $N$ points every epoch, producing $2{-}19\times$ slowdowns that grow with dataset size. On covtype ($N{=}581\text{k}$, $k{=}32$): $18.9\times$ slower; on SUSY ($N{=}5\text{M}$, $k{=}32$): $9.6\times$ slower. On large datasets most points are stable after early epochs, and screening correctly identifies them.

\paragraph{The parallel path is critical for large $k$ and large $N$.}
Table~\ref{tab:ablation-parallel} shows the fraction of non-screened point operations resolved by the parallel BLAS path versus those requiring sequential $O(k)$ processing.

\begin{table}[!h]
\centering
\caption{Impact of removing the parallel path. Without it, all non-screened operations become sequential $O(k)$-per-point work, growing dramatically with $k$ and $N$ --- up to $1{,}671\times$ on SUSY ($N{=}5\text{M}$).}
\label{tab:ablation-parallel}
\small
\begin{tabular}{llrrr}
\toprule
Dataset & $k$ & \% Parallel (full) & Seq.\ ops ratio & Iter.\ ratio \\
\midrule
phishing     &   8 &  69.7\% &      3.5$\times$ & 1.04$\times$ \\
             &  32 &  71.5\% &      3.2$\times$ & 0.82$\times$ \\
\midrule
Letter       &   8 &  80.4\% &      5.2$\times$ & 1.00$\times$ \\
             &  32 &  86.4\% &      7.4$\times$ & 0.95$\times$ \\
             & 128 &  90.5\% &     11.3$\times$ & 1.13$\times$ \\
\midrule
mnist        &   8 &  87.7\% &      8.3$\times$ & 1.00$\times$ \\
             &  32 &  89.9\% &     10.3$\times$ & 0.98$\times$ \\
             & 128 &  94.4\% &     23.1$\times$ & 1.23$\times$ \\
\midrule
connect-4    &   8 &  81.0\% &      5.3$\times$ & 0.96$\times$ \\
             &  32 &  85.1\% &      6.9$\times$ & 0.99$\times$ \\
             & 128 &  87.1\% &      6.6$\times$ & 0.79$\times$ \\
\midrule
aloi         &   8 &  94.5\% &     19.5$\times$ & 1.02$\times$ \\
             &  32 &  95.5\% &     23.7$\times$ & 1.04$\times$ \\
             & 128 &  96.8\% &     24.5$\times$ & 0.61$\times$ \\
\midrule
covtype      &   8 &  98.8\% &     90.1$\times$ & 1.07$\times$ \\
             &  32 &  99.1\% &    121.5$\times$ & 5.83$\times$ \\
             & 128 &  99.2\% &    125.6$\times$ & 0.16$\times$ \\
\midrule
SUSY         &   8 &  99.8\% &    604.2$\times$ & 0.92$\times$ \\
             &  32 &  99.9\% &  1,039.9$\times$ & 0.92$\times$ \\
             & 128 &  99.9\% &  1,671.3$\times$ & 0.87$\times$ \\
\bottomrule
\end{tabular}
\end{table}

Without the parallel path, the volume of sequential operations increases by orders of magnitude on large datasets: $90{-}130\times$ on covtype ($N{=}581\text{k}$) and $604{-}1{,}671\times$ on SUSY ($N{=}5\text{M}$). The growth with $k$ is clear: at $k{=}128$ the full algorithm resolves ${\geq}99\%$ of non-screened points via bulk BLAS, but removing this forces all of them into per-point $O(k)$ sequential processing. On SUSY $k{=}32$ this translates directly to a $1.27\times$ wall-time slowdown even in single-threaded execution.

\paragraph{Observed pruning rates.}
After epoch~2, cross-batch screening removes 50--95\% of points.  Of
the remainder, the parallel safety path certifies 70--99.9\%
(\autoref{tab:ablation-parallel}); the sequential fallback handles
fewer than 5\% of points in steady state.

\paragraph{Fixed small batch size is catastrophic.}
Setting $B{=}B_{\min}{=}64$ produces $1.4{-}9\times$ slowdowns across all configurations due to excessive per-batch overhead. The largest impact is on SUSY ($8.9\times$ at $k{=}32$). The adaptive EMA-based sizing avoids this failure mode automatically.

\paragraph{The sequential drift-aware bounds provide modest but consistent benefit.}
Removing the sequential kernel's S1/S2 bounds (brute-forcing ambiguous points) incurs $1.3{-}1.5\times$ overhead on high-dimensional data (mnist $D{=}780$), where recomputing exact distances is expensive. On lower-dimensional data the brute-force cost is negligible.

\subsection{Hyperparameter Sensitivity}\label{app:ablation-params}

We vary each hyperparameter independently by $10\times$ above and below its default, holding all others constant. Table~\ref{tab:ablation-params} reports the wall-time ratio relative to the default configuration.

\begin{table}[!h]
\centering
\caption{Hyperparameter sensitivity: wall-time ratio vs.\ default. Values near 1.0 indicate insensitivity. All configurations produce equivalent inertia (${<}0.01\%$ variation).}
\label{tab:ablation-params}
\small
\begin{tabular}{lrrccc}
\toprule
Parameter & Default & Swept & \multicolumn{3}{c}{Median time ratio (range across datasets)} \\
\cmidrule(lr){4-6}
 & & values & $0.1\times$ & $1\times$ & $10\times$ \\
\midrule
$B_{\min}$ & 64 & 6, 64, 640 & 1.00 (0.99--1.02) & 1.00 & 1.00 (0.99--1.01) \\
$B_{\max}$ & 4096 & 410, 4096, 40960 & 1.81 (0.65--2.75) & 1.00 & 1.08 (0.72--3.29) \\
$\lambda$  & 0.1 & 0.01, 0.1, 1.0 & 1.00 (0.91--1.25) & 1.00 & 1.00 (0.96--1.02) \\
$c$        & 0.5 & 0.05, 0.5, 5.0 & 1.00 (0.82--1.44) & 1.00 & 1.00 (0.97--1.02) \\
\bottomrule
\end{tabular}
\end{table}

Only $B_{\max}$ has a meaningful impact: reducing it $10\times$ to 410 causes median $1.8\times$ slowdown (more batches per epoch), while increasing it $10\times$ to 40,960 produces mixed results---faster on large-$N$/low-$D$ datasets but up to $3.3\times$ slower on high-$D$ data (mnist $k{=}128$) due to oversized GEMM operations. The remaining parameters ($B_{\min}$, $\lambda$, $c$) are essentially inert: $10\times$ perturbations in either direction produce ${<}5\%$ median runtime variation. Combined with zero effect on solution quality, this confirms that users need not tune these parameters.

%% file: sections/appendix_proofs.tex
\section{Supporting Lemmas and Proofs}\label{app:proofs}

The following lemma is used throughout the proofs below.

\begin{lemma}[Center drift]\label{lem:center-drift}
Suppose point $\mathbf{x}_i$ is reassigned from cluster $C_m$
(with $n_m \geq 2$) to cluster $C_{j^{\star}}$.  Let
$\boldsymbol{\mu}_m, \boldsymbol{\mu}_{j^{\star}}$ denote the
centroids \emph{before} the move.  Then the centroids after the
move satisfy $\boldsymbol{\mu}_m'
  = \frac{n_m \, \boldsymbol{\mu}_m - \mathbf{x}_i}{n_m - 1}$ and $\boldsymbol{\mu}_{j^{\star}}'
  = \frac{n_{j^{\star}} \boldsymbol{\mu}_{j^{\star}}
            + \mathbf{x}_i}{n_{j^{\star}} + 1}$ with changes $\|\boldsymbol{\mu}_m' - \boldsymbol{\mu}_m\|
  = \frac{\|\mathbf{x}_i - \boldsymbol{\mu}_m\|}{n_m - 1}$ and $\|\boldsymbol{\mu}_{j^{\star}}'
    - \boldsymbol{\mu}_{j^{\star}}\|
  = \frac{\|\mathbf{x}_i
            - \boldsymbol{\mu}_{j^{\star}}\|}{n_{j^{\star}} + 1}$ respectively.
\end{lemma}

\begin{proof}[Proof of Lemma~\ref{lem:center-drift} (Center drift)]
For removal:
\[
  \boldsymbol{\mu}_m' - \boldsymbol{\mu}_m
  = \frac{n_m \boldsymbol{\mu}_m - \mathbf{x}_i}{n_m - 1}
    - \boldsymbol{\mu}_m
  = \frac{\boldsymbol{\mu}_m - \mathbf{x}_i}{n_m - 1}.
\]
Taking norms gives
the removal result.
For insertion:
\[
  \boldsymbol{\mu}_{j^{\star}}' - \boldsymbol{\mu}_{j^{\star}}
  = \frac{n_{j^{\star}} \boldsymbol{\mu}_{j^{\star}}
          + \mathbf{x}_i}{n_{j^{\star}} + 1}
    - \boldsymbol{\mu}_{j^{\star}}
  = \frac{\mathbf{x}_i
          - \boldsymbol{\mu}_{j^{\star}}}{n_{j^{\star}} + 1},
\]
giving the result of the Lemma. 
\end{proof}

\begin{proof}[Proof of Lemma~\ref{lem:commutative} (Commutativity of centroid updates)]
Each incremental update
$\boldsymbol{\mu} \gets (n\boldsymbol{\mu} \pm \mathbf{x}_i)
                        / (n \pm 1)$
preserves the invariant
$\boldsymbol{\mu} = (\sum_{i \in C_j} \mathbf{x}_i) / |C_j|$.
After any permutation of the $|S^{-}| + |S^{+}|$ operations,
the membership set is
$C_j' = (C_j \setminus S^{-}) \cup S^{+}$,
whose member sum and cardinality are permutation-invariant.
\end{proof}

\begin{proof}[Proof of Lemma~\ref{lem:dist-bounds} (Distance bound propagation)]
\textup{(i)}~By the triangle inequality,
$\|\mathbf{x}_b - \boldsymbol{\mu}_j^{(b-1)}\|
 \leq \|\mathbf{x}_b - \boldsymbol{\mu}_j^{(0)}\|
      + \|\boldsymbol{\mu}_j^{(b-1)}
          - \boldsymbol{\mu}_j^{(0)}\|
 \leq \sqrt{\hat{d}_{bj}} + \delta_j^{(b-1)}$,
and the reverse triangle inequality gives the lower bound.
Squaring preserves order on $[0,\infty)$.
\textup{(ii)}~The Hartigan weights
$n_j^{(b-1)}/(n_j^{(b-1)} \pm 1)$ are positive, so
multiplying
by the appropriate weight
preserves the inequalities.  
\end{proof}

\begin{theorem}[Correctness of Batched Hartigan]%
\label{thm:batched-correct}
Consider standard Hartigan
(Algorithm~\ref{alg:hartigan}) processing points in a fixed order
$1, 2, \dots, N$.  Let Batched Hartigan process the same
points in the same order, partitioned into contiguous batches of
size at most $B$.  If Batched Hartigan makes a reassignment only
under condition~\textup{(S1)} and skips a reassignment only
under condition~\textup{(S2)}, falling back to exact distance
computation in all ambiguous cases, then the sequence of moves
produced by Batched Hartigan is identical to that of standard
Hartigan.
\end{theorem}

\begin{proof}[Proof of Theorem~\ref{thm:batched-correct} (Correctness of Batched Hartigan)]
By induction on the position in the processing order.

\emph{Base case.}\;
The first point in the first batch has
$\delta_j^{(0)} = 0$ for all~$j$, so all bounds are exact and
conditions (S1), (S2) reduce to the standard Hartigan comparison.

\emph{Inductive step.}\;
Suppose that through points $1, \dots, b{-}1$, Batched
Hartigan has produced the same moves as standard Hartigan.  Then
the cluster state (memberships, sizes, centroids) is identical at
step~$b$.  For point~$b$:
\begin{itemize}
  \item The precomputed distances $\hat{d}_{bj}$ were computed
        with respect to $\boldsymbol{\mu}_j^{(0)}$.
  \item The cumulative drifts $\delta_j^{(b-1)}$ correctly bound
        $\|\boldsymbol{\mu}_j^{(b-1)}
           - \boldsymbol{\mu}_j^{(0)}\|$
        by Definition~\ref{def:cum-drift}.
  \item The sizes $n_j^{(b-1)}$ are tracked exactly.
\end{itemize}
If (S1) holds, the true Hartigan cost to $j^{\star}$ is at most
$H_{j^{\star}}^{\mathrm{hi}}$, which is strictly less than the
lower bound on every other option.  Exact Hartigan would also
choose $j^{\star}$.
If (S2) holds, the true cost to every candidate is at least
$H_j^{\mathrm{lo}} \geq H_m^{\mathrm{hi}}$, which exceeds the
true cost of staying.  Exact Hartigan would also keep the point.
If neither holds, exact distances are recomputed, so the decision
is trivially identical.  In every case the move matches, the
induction carries, and the cluster state remains synchronised.
\end{proof}

\begin{proof}[Proof of Lemma~\ref{lem:aggregate-bounds} (Aggregate batch bounds)]
\textup{(i)}\;
Let $D_j^{(s)} := \|\boldsymbol{\mu}_j^{(s)} -
\boldsymbol{\mu}_j^{(0)}\|$ denote the total drift of
cluster~$j$ after $s$~moves within the batch have been processed
(in any order), with $D_j^{(0)} = 0$.  We show
$D_j^{(s)} \leq \delta_j^{\mathrm{agg}}$ for all~$s$.

\emph{Per-move drift.}\;
Suppose that at step~$s$ a point $\mathbf{x}_{b_s}$ is added to
or removed from cluster~$j$.  By
Lemma~\ref{lem:center-drift}, the centroid shift is
\begin{equation}\label{eq:per-move-drift}
  \|\boldsymbol{\mu}_j^{(s)} - \boldsymbol{\mu}_j^{(s-1)}\|
  \;=\;
  \frac{\|\mathbf{x}_{b_s} - \boldsymbol{\mu}_j^{(s-1)}\|}
       {n_j^{(s-1)} \pm 1} \,,
\end{equation}
where $n_j^{(s-1)} \pm 1$ denotes $n_j^{(s-1)} - 1$ for removal
or $n_j^{(s-1)} + 1$ for insertion (both $\geq n_j^{\min}$
since at most $L_j$ points leave cluster~$j$).

\emph{Bounding the intermediate distance.}\;
The numerator involves the intermediate centroid
$\boldsymbol{\mu}_j^{(s-1)}$, not the precomputed
$\boldsymbol{\mu}_j^{(0)}$.  By the triangle inequality:
\begin{equation}\label{eq:intermediate-dist-bound}
  \|\mathbf{x}_{b_s} - \boldsymbol{\mu}_j^{(s-1)}\|
  \;\leq\;
  \|\mathbf{x}_{b_s} - \boldsymbol{\mu}_j^{(0)}\|
  \;+\;
  \|\boldsymbol{\mu}_j^{(s-1)} - \boldsymbol{\mu}_j^{(0)}\|
  \;=\;
  \sqrt{\hat{d}_{b_s,j}} + D_j^{(s-1)}.
\end{equation}

\emph{Recurrence.}\;
Combining~\eqref{eq:per-move-drift}
and~\eqref{eq:intermediate-dist-bound}, and bounding the
denominator below by $n_j^{\min}$:
\begin{equation}\label{eq:drift-recurrence}
  D_j^{(s)}
  \;\leq\;
  D_j^{(s-1)}
  + \frac{\sqrt{\hat{d}_{b_s,j}} + D_j^{(s-1)}}{n_j^{\min}}
  \;=\;
  D_j^{(s-1)}\!\Bigl(1 + \frac{1}{n_j^{\min}}\Bigr)
  + \frac{\sqrt{\hat{d}_{b_s,j}}}{n_j^{\min}} \,.
\end{equation}
Write $\alpha = 1/n_j^{\min}$.  With $D_j^{(0)} = 0$ the
recurrence
$D^{(s)} \leq D^{(s-1)}(1+\alpha)
 + \alpha\,\sqrt{\hat{d}_{b_s,j}}$
unrolls by induction to
\begin{equation}\label{eq:drift-unrolled}
  D_j^{(M_j)}
  \;\leq\;
  \alpha
  \sum_{t=1}^{M_j} \sqrt{\hat{d}_{b_t,j}}
    \,(1+\alpha)^{M_j - t}
  \;\leq\;
  \alpha\,(1+\alpha)^{M_j - 1}
  \cdot R_j \,,
\end{equation}
where the second inequality
holds because $(1+\alpha)^{M_j - t}$ is largest for $t = 1$
and $R_j = \sum_{t} \sqrt{\hat{d}_{b_t,j}}$
(Lemma~\ref{lem:aggregate-bounds}).

It remains to bound $\alpha(1+\alpha)^{M_j-1}$ by
$1/(n_j^{\min} - M_j)$.  We use the elementary inequality
\begin{equation}\label{eq:exp-bound}
  \bigl(1 + \tfrac{1}{n}\bigr)^{M}
  \;\leq\;
  e^{M/n}
  \;\leq\;
  \frac{n}{n - M}
  \qquad (0 \leq M < n),
\end{equation}
where the first step is $(1+1/n)^n \leq e$ and the second
follows from $e^{x}(1-x) \leq 1$ for $x \geq 0$ (since
$\frac{d}{dx}[e^x(1-x)] = -x\,e^x \leq 0$
and the value at $x=0$ is~$1$).  Applying~\eqref{eq:exp-bound}
to~\eqref{eq:drift-unrolled} with $n = n_j^{\min}$ and
$M = M_j$:
\[
  D_j^{(M_j)}
  \;\leq\;
  \frac{1}{n_j^{\min}} \cdot
  \frac{n_j^{\min}}{n_j^{\min} - M_j}
  \cdot R_j
  \;=\;
  \frac{R_j}{n_j^{\min} - M_j}
  \;=\;
  \delta_j^{\mathrm{agg}}.
\]
Since only $s \leq M_j$ of the batch moves touch cluster~$j$
and the recurrence is monotone in~$s$, the bound
$D_j^{(s)} \leq \delta_j^{\mathrm{agg}}$ holds for every
intermediate step.

\textup{(ii)}\;
During sequential processing, the size of cluster~$j$ lies in
$[n_j - L_j,\; n_j + J_j]$.  The Hartigan weight for the
current cluster is $n_m^{(s)} / (n_m^{(s)} - 1)$, which is
maximised when $n_m^{(s)}$ is smallest.  The smallest possible
size when processing point~$b$ is $n_m - L_m^{-b}$, where
$L_m^{-b}$ excludes $b$'s own tentative departure (which has not
yet occurred when $b$ is evaluated),
giving $w_m^{\max}$ (Lemma~\ref{lem:aggregate-bounds}(ii)).  The candidate weight
$n_j^{(s)}/(n_j^{(s)}+1)$ is minimised when $n_j^{(s)}$ is
smallest, giving $w_j^{\min}$.  Combining these
weights with the squared-distance bounds
$\bigl[\sqrt{\hat{d}_{bj}} \pm
\delta_j^{\mathrm{agg}}\bigr]_+^{2}$
from~(i) yields the Hartigan cost bounds.

\textup{(iii)}\;
For safe-move verification, the worst-case directions reverse: we
need the \emph{smallest} current-cluster weight (maximise
$n_m^{(s)}$, giving $n_m + J_m$) and the \emph{largest}
candidate weight (maximise $n_j^{(s)}$, giving $n_j + J_j$).
\end{proof}

\begin{proof}[Proof of Lemma~\ref{lem:generalized-drift} (Aggregate bounds with pre-existing drift)]
The argument follows the drift
recurrence~\eqref{eq:drift-recurrence} from
Lemma~\ref{lem:aggregate-bounds}, but with a nonzero initial
condition.  Write $\alpha = 1/n_j^{\min}$ and let $D_j^{(s)}$
denote the total drift from the GEMM centroids after $s$~moves
within $\mathcal{A}$ have been processed.  Now
$D_j^{(0)} = \Delta_j^{\mathrm{pre}}$ (not zero).

The per-move drift bound~\eqref{eq:per-move-drift} and
intermediate distance bound~\eqref{eq:intermediate-dist-bound}
are unchanged, since they depend only on the triangle inequality
and the current cluster size (which is at least $n_j^{\min}$).
The recurrence~\eqref{eq:drift-recurrence} therefore reads
\[
  D_j^{(s)}
  \;\leq\;
  D_j^{(s-1)} (1 + \alpha)
  + \alpha\, \sqrt{\hat{d}_{b_s,j}} \,,
\]
with $D_j^{(0)} = \Delta_j^{\mathrm{pre}}$.
Unrolling:
\begin{equation}\label{eq:drift-unrolled-gen}
  D_j^{(M_j)}
  \;\leq\;
  \Delta_j^{\mathrm{pre}}\, (1 + \alpha)^{M_j}
  \;+\;
  \alpha\,(1 + \alpha)^{M_j - 1} \cdot R_j^{\mathcal{A}}.
\end{equation}
Applying the elementary bound
$(1 + 1/n)^{M} \leq n/(n-M)$
from~\eqref{eq:exp-bound} with $n = n_j^{\min}$ and $M = M_j$:
\begin{align*}
  D_j^{(M_j)}
  &\;\leq\;
  \Delta_j^{\mathrm{pre}}
  \cdot \frac{n_j^{\min}}{n_j^{\min} - M_j}
  \;+\;
  \frac{R_j^{\mathcal{A}}}{n_j^{\min} - M_j}
  \\[4pt]
  &\;=\;
  \frac{\Delta_j^{\mathrm{pre}}\cdot n_j^{\min}
        + R_j^{\mathcal{A}}}
       {n_j^{\min} - M_j}
  \;=\; \delta_j^{\mathrm{eff}}.
\end{align*}
Since $D_j^{(s)}$ is monotonically non-decreasing in~$s$ and only
$M_j$ moves touch cluster~$j$, the bound holds at every
intermediate step.  When $\Delta_j^{\mathrm{pre}} = 0$, the
expression reduces to
$\delta_j^{\mathrm{agg}} = R_j / (n_j^{\min} - M_j)$
from Lemma~\ref{lem:aggregate-bounds}.
\end{proof}

\begin{lemma}[Aggregate bounds with pre-existing drift]%
\label{lem:generalized-drift}
Let $\mathcal{A}$ be a set of points with tentative decisions
having aggregate flow $L_j^{\mathcal{A}}, J_j^{\mathcal{A}}$ and
batch quantities as in Lemma~\ref{lem:aggregate-bounds}, computed
with respect to sizes~$n_j$.
Suppose
$\|\boldsymbol{\mu}_j^{(\mathrm{current})}
   - \boldsymbol{\mu}_j^{(\mathrm{GEMM})}\|
\leq \Delta_j^{\mathrm{pre}}$.
Write $n_j^{\min} = n_j - L_j^{\mathcal{A}}$,
$M_j = L_j^{\mathcal{A}} + J_j^{\mathcal{A}}$, and assume
$M_j < n_j^{\min}$.  Then at every intermediate step,
$\|\boldsymbol{\mu}_j^{(\mathrm{step})}
   - \boldsymbol{\mu}_j^{(\mathrm{GEMM})}\|
\leq \delta_j^{\mathrm{eff}}
:= (\Delta_j^{\mathrm{pre}} \cdot n_j^{\min} + R_j^{\mathcal{A}})
   / (n_j^{\min} - M_j)$,
where
$R_j^{\mathcal{A}}
 = \sum_{b \in \mathcal{A}:\,T(b)\text{ touches }j}
   \sqrt{\hat{d}_{bj}}$.
\end{lemma}

\begin{proof}[Proof of Theorem~\ref{thm:partial-parallel} (Partial parallel batch correctness)]
\textup{(a)}\;
Restricting the batch to $\mathcal{S}$ gives reduced flow
$L_j^{\mathcal{S}} \leq L_j$,
$J_j^{\mathcal{S}} \leq J_j$,
$R_j^{\mathcal{S}} \leq R_j$, and
$n_j^{\min,\mathcal{S}} = n_j - L_j^{\mathcal{S}}
 \geq n_j^{\min}$.  The restricted drift bound therefore
satisfies
\[
  \delta_j^{\mathcal{S}}
  \;=\;
  \frac{R_j^{\mathcal{S}}}
       {n_j^{\min,\mathcal{S}} - M_j^{\mathcal{S}}}
  \;\leq\;
  \delta_j^{\mathrm{agg}}.
\]
Points in $\mathcal{S}$ satisfy (S1) or (S2) under the larger
bound $\delta_j^{\mathrm{agg}}$, hence they also satisfy the
conditions under $\delta_j^{\mathcal{S}}$.  By
Lemma~\ref{lem:aggregate-bounds}, the drift at every
intermediate step of a hypothetical sequential processing of
$\mathcal{S}$ remains within
$\delta_j^{\mathcal{S}}
 \leq \delta_j^{\mathrm{agg}}$, so each decision is correct:
the Hartigan cost strictly decreases.  By
Lemma~\ref{lem:commutative}, the bulk centroid update
yields the same final state as any sequential ordering of
$\mathcal{S}$'s movers.

\textup{(b)}\;
After executing $\mathcal{S}$'s bulk update, the actual
per-cluster displacement from the GEMM reference is
$\Delta_j
 = \|\boldsymbol{\mu}_j^{(\text{after } \mathcal{S})}
     - \boldsymbol{\mu}_j^{(\mathrm{GEMM})}\|$.
This is an \emph{exact} measurement (not a bound).  Sequential
processing of~$\mathcal{A}$ starting from this centroid state
uses the SequentialBatch algorithm
(Algorithm~\ref{alg:sequential-batch}) with per-cluster drift
initialised to~$\Delta_j$.  The drift at every subsequent step
is bounded by Lemma~\ref{lem:generalized-drift}
(Lemma~\ref{lem:generalized-drift}).  Each point in~$\mathcal{A}$
therefore evaluates Hartigan cost bounds that are valid---tighter
than under $\delta_j^{\mathrm{agg}}$---and makes the optimal
reassignment decision or falls through to the exact-distance
computation.

\textup{(c)}\;
Every accepted move (whether from $\mathcal{S}$ or
$\mathcal{A}$) strictly decreases $\mathcal{L}(\mathcal{C})$:
moves from $\mathcal{S}$ by part~(a), and moves from
$\mathcal{A}$ because each sequential decision is a valid
Hartigan move with respect to the current centroid state.
Since $\mathcal{L}$ is bounded below and strictly decreases with
each accepted move, the algorithm terminates in a finite number of
epochs.

At convergence, no points move, so $L_j = J_j = M_j = 0$ and
$\delta_j^{\mathrm{agg}} = 0$ for all~$j$.  Every point is in
$\mathcal{S}$ (all are safe stayers) and the partition satisfies:
for every point $\mathbf{x}_i$ in cluster~$m$, no single-point
move to any cluster~$j \neq m$ would decrease
$\mathcal{L}(\mathcal{C})$.  This is exactly the Hartigan
fixed-point condition.
\end{proof}

\begin{proof}[Proof of Lemma~\ref{lem:cross-batch} (Cross-batch bound validity)]
Let the bounds $u_i, \ell_i$ be set exactly after batch~$t_0$
from a GEMM, with snapshots $\sigma_i^{u} = \sigma_{m(i)}^{(t_0)}$
and $\Sigma_i^{\ell} = \Sigma_{\max}^{(t_0)}$.

\emph{Upper bound.}\;
By the triangle inequality, at any later batch~$t > t_0$:
\[
  \|\mathbf{x}_i - \boldsymbol{\mu}_{m(i)}^{(t)}\|
  \;\leq\;
  \|\mathbf{x}_i - \boldsymbol{\mu}_{m(i)}^{(t_0)}\|
  + \|\boldsymbol{\mu}_{m(i)}^{(t)}
      - \boldsymbol{\mu}_{m(i)}^{(t_0)}\|
  \;\leq\;
  u_i + \bigl(\sigma_{m(i)}^{(t)} - \sigma_{m(i)}^{(t_0)}\bigr)
  = u_i^{\mathrm{eff}},
\]
where the second inequality uses the triangle inequality on the
cumulative drift:
$\|\boldsymbol{\mu}_{m(i)}^{(t)} -
   \boldsymbol{\mu}_{m(i)}^{(t_0)}\|
 \leq \sum_{s=t_0+1}^{t} \delta_{m(i)}^{(s)}
 = \sigma_{m(i)}^{(t)} - \sigma_{m(i)}^{(t_0)}$.

\emph{Lower bound.}\;
For any cluster $j \neq m(i)$, the reverse triangle inequality
gives
$\|\mathbf{x}_i - \boldsymbol{\mu}_j^{(t)}\|
 \geq \|\mathbf{x}_i - \boldsymbol{\mu}_j^{(t_0)}\|
      - (\sigma_j^{(t)} - \sigma_j^{(t_0)})$.
Taking the minimum over $j \neq m(i)$:
\[
  \min_{j \neq m(i)}
  \|\mathbf{x}_i - \boldsymbol{\mu}_j^{(t)}\|
  \;\geq\;
  \ell_i
  - \max_{j \neq m(i)}
    \bigl(\sigma_j^{(t)} - \sigma_j^{(t_0)}\bigr).
\]
It remains to show
$\Sigma_{\max}^{(t)} - \Sigma_{\max}^{(t_0)}
 \geq \max_j (\sigma_j^{(t)} - \sigma_j^{(t_0)})$.
Expanding the left side:
\[
  \Sigma_{\max}^{(t)} - \Sigma_{\max}^{(t_0)}
  \;=\;
  \sum_{s=t_0+1}^{t} \max_j\, \delta_j^{(s)}
  \;\geq\;
  \max_j \sum_{s=t_0+1}^{t} \delta_j^{(s)}
  \;=\;
  \max_j
  \bigl(\sigma_j^{(t)} - \sigma_j^{(t_0)}\bigr),
\]
where the inequality holds because
$\sum_s \max_j f_j(s) \geq \max_j \sum_s f_j(s)$
for any non-negative $f_j(s)$ (the sum of per-step maxima
dominates the maximum of per-cluster sums).
Therefore
$\ell_i^{\mathrm{eff}} = \ell_i -
 (\Sigma_{\max}^{(t)} - \Sigma_i^{\ell})$
is a valid lower bound.
\end{proof}

%% file: sections/appendix_speedup_figure.tex
\section{GPU Speedup}

\begin{figure}[!h]
\centering
\begin{tikzpicture}
\begin{axis}[
  width=\columnwidth,
  height=0.35\columnwidth,
  ybar,
  bar width=10pt,
  symbolic x coords={2, 4, 8, 16, 32, 64, 128},
  xtick=data,
  x tick label style={font=\small},
  xlabel={Number of clusters ($k$)},
  ylabel={Speedup ($\times$)},
  ymin=0, ymax=3.5,
  enlarge x limits=0.08,
  legend style={at={(0.02,0.97)}, anchor=north west, font=\scriptsize,
    draw=none,
    fill=white,
    fill opacity=0.8,
    area legend
  },
  ymajorgrids=true,
  grid style={dashed, gray!40},
]

\addplot[
  forget plot,
  draw=none,
  error bars/.cd,
  y dir=both,
  y explicit,
  error bar style={thick, black, line width=0.8pt}
] coordinates {
  (2, 0.807) +- (0, 0.351)
  (4, 0.826) +- (0, 0.466)
  (8, 0.778) +- (0, 0.511)
  (16, 0.814) +- (0, 0.589)
  (32, 0.830) +- (0, 0.593)
  (64, 0.990) +- (0, 1.019)
  (128, 1.543) +- (0, 1.686)
};

\addplot[fill=Set1-A, draw=none] coordinates {
  (2, 0.807)
  (4, 0.826)
  (8, 0.778)
  (16, 0.814)
  (32, 0.830)
  (64, 0.990)
  (128, 1.543)
};

\addlegendentry{Optimized / Baseline}

\draw[dashed, gray, thick]
  (axis cs:2,1.0) -- (axis cs:128,1.0);

\node[font=\scriptsize, anchor=south] at (axis cs:64, 2.2) {0.99$\times$};
\node[font=\scriptsize, anchor=south] at (axis cs:128, 3.4) {1.54$\times$};

\end{axis}
\end{tikzpicture}
\caption{%
\textbf{GPU speedup after optimization (NVIDIA Tesla T4).}
Bars show mean speedup across 14 datasets (Letter Recognition, MAGIC Gamma Telescope, SUSY, Sepsis Survival, Skin Segmentation, Statlog Shuttle, Superconductivity, aloi, cod-rna, connect-4, covtype, mnist, phishing, skin-nonskin).
Error bars indicate min-max range across datasets.
\textbf{Optimizations:}
(1) Persistent GPU memory: centers and cluster sizes remain on GPU across batches within each epoch, updated in-place;
(2) GPU-resident distance computation: GEMM output ($B \times k$) stays on GPU for downstream operations;
(3) Vectorized tentative decisions: broadcast Hartigan weights to $(B \times k)$, compute element-wise costs, parallel row-wise argmin on GPU instead of nested CPU loops;
(4) GPU-accelerated S1/S2 safety checks: compute distance bounds on GPU, transfer only boolean mask ($B$ bits) instead of intermediate matrices ($2Bk$ floats).
\textbf{Performance trends:}
For $k \le 32$, fixed GPU overheads (kernel launch, synchronization, memory allocation) dominate, resulting in 0.81$\times$ average speedup (19\% slower).
For $k = 64$, performance approaches parity (0.99$\times$).
For $k = 128$, the $O(Bk)$ operations achieve high GPU occupancy and amortize fixed costs, yielding 1.54$\times$ average speedup with best case 3.2$\times$ on SUSY ($N=5$M, $D=18$).
See Figure~\ref{fig:gpu-speedup-per-dataset} for per-dataset breakdown.
}%
\label{fig:gpu-speedup}
\end{figure}
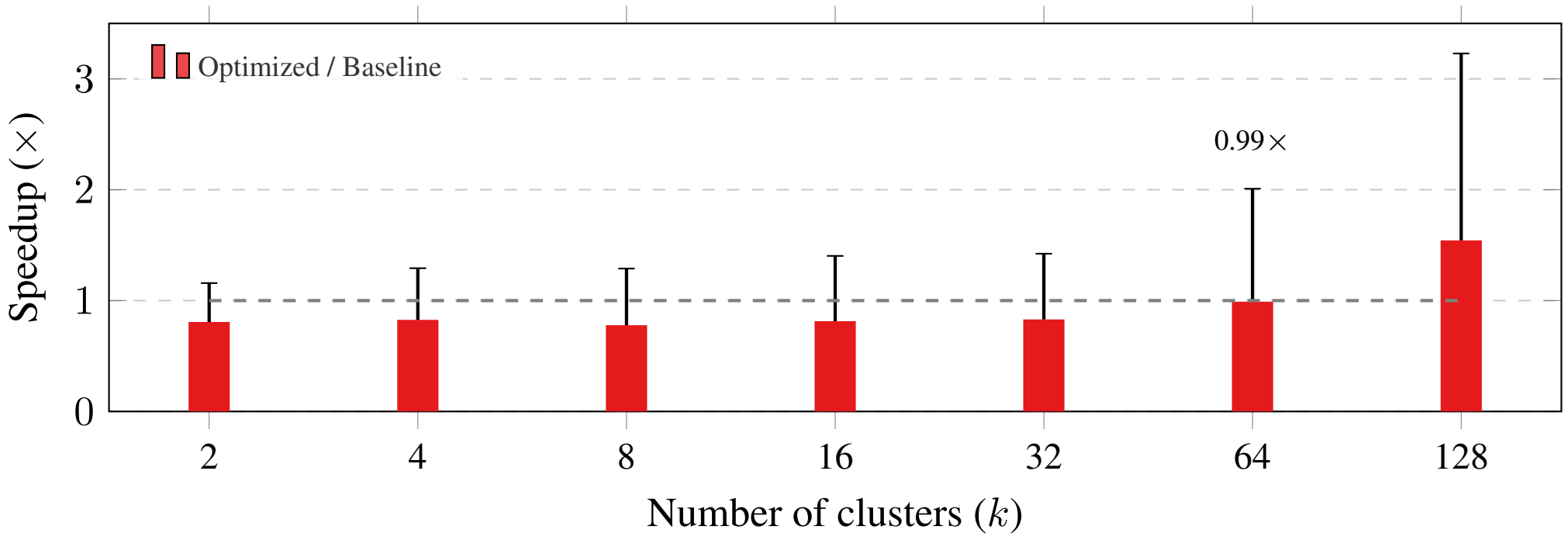

\begin{figure}[!h]
\centering
\begin{tikzpicture}
\begin{axis}[
  width=\columnwidth,
  height=0.5\columnwidth,
  symbolic x coords={2, 4, 8, 16, 32, 64, 128},
  xtick=data,
  xlabel={Number of clusters ($k$)},
  ylabel={Speedup ($\times$)},
  ymin=0, ymax=3.6,
  legend style={
    at={(0.02,0.97)},
    anchor=north west,
    font=\tiny,
    legend columns=2,
    draw=none,
    fill=white,
    fill opacity=0.8
  },
  ymajorgrids=true,
  grid style={dashed, gray!40},
]

\addplot[
  color=Set1-A,
  mark=*,
  line width=1.5pt,
  mark size=2pt
] coordinates {
  (2, 1.091) (4, 1.154) (8, 1.200) (16, 1.286)
  (32, 0.766) (64, 2.009) (128, 3.229)
};
\addlegendentry{SUSY}

\addplot[
  color=Set1-B,
  mark=square*,
  line width=1.5pt,
  mark size=2pt
] coordinates {
  (2, 1.000) (4, 1.023) (8, 1.089) (16, 0.994)
  (32, 1.131) (64, 1.514) (128, 2.169)
};
\addlegendentry{covtype}

\addplot[
  color=Set1-C,
  mark=triangle*,
  line width=1.5pt,
  mark size=2.5pt
] coordinates {
  (2, 1.060) (4, 1.016) (8, 1.128) (16, 1.107)
  (32, 1.303) (64, 0.903) (128, 2.673)
};
\addlegendentry{Skin Seg}

\addplot[
  color=Set1-D,
  mark=diamond*,
  line width=1.5pt,
  mark size=2.5pt
] coordinates {
  (2, 0.510) (4, 0.491) (8, 0.540) (16, 0.621)
  (32, 0.673) (64, 0.984) (128, 1.442)
};
\addlegendentry{Letter}

\addplot[
  color=Set1-E,
  mark=pentagon*,
  line width=1.5pt,
  mark size=2.5pt
] coordinates {
  (2, 1.010) (4, 1.063) (8, 0.267) (16, 0.225)
  (32, 0.237) (64, 0.287) (128, 0.459)
};
\addlegendentry{Shuttle}

\addplot[
  color=Set1-F,
  mark=otimes*,
  line width=1.5pt,
  mark size=2pt
] coordinates {
  (2, 0.456) (4, 0.360) (8, 0.405) (16, 0.379)
  (32, 0.438) (64, 0.686) (128, 0.903)
};
\addlegendentry{MAGIC}

\draw[dashed, gray, thick]
  (axis cs:2,1.0) -- (axis cs:128,1.0);

\end{axis}
\end{tikzpicture}
\caption{%
\textbf{Per-dataset GPU speedup breakdown (line plot).}
Six representative datasets shown across all $k$ values.
\textbf{Key observations:}
(1) Large datasets with high dimensionality benefit most: SUSY ($N=5$M, $D=18$) achieves 3.2$\times$ at $k=128$; covtype ($N=581$K, $D=54$) achieves 2.2$\times$; Skin Segmentation ($N=245$K, $D=3$) achieves 2.7$\times$.
(2) Small, low-dimensional datasets suffer: MAGIC Gamma ($N=19$K, $D=10$) and Letter Recognition ($N=20$K, $D=16$) show 50--60\% slowdown for $k \le 32$.
(3) Pathological case: Statlog Shuttle ($N=58$K, $D=7$) shows severe degradation (0.23$\times$ at $k=16$), likely due to small batch sizes and poor GPU occupancy.
(4) Crossover point: Most datasets achieve parity around $k=64$ and benefit at $k=128$.
This heterogeneity suggests \textbf{adaptive backend selection}: use CPU for small $k$ or small datasets, GPU for large $k$ with sufficient data.
}%
\label{fig:gpu-speedup-per-dataset}
\end{figure}
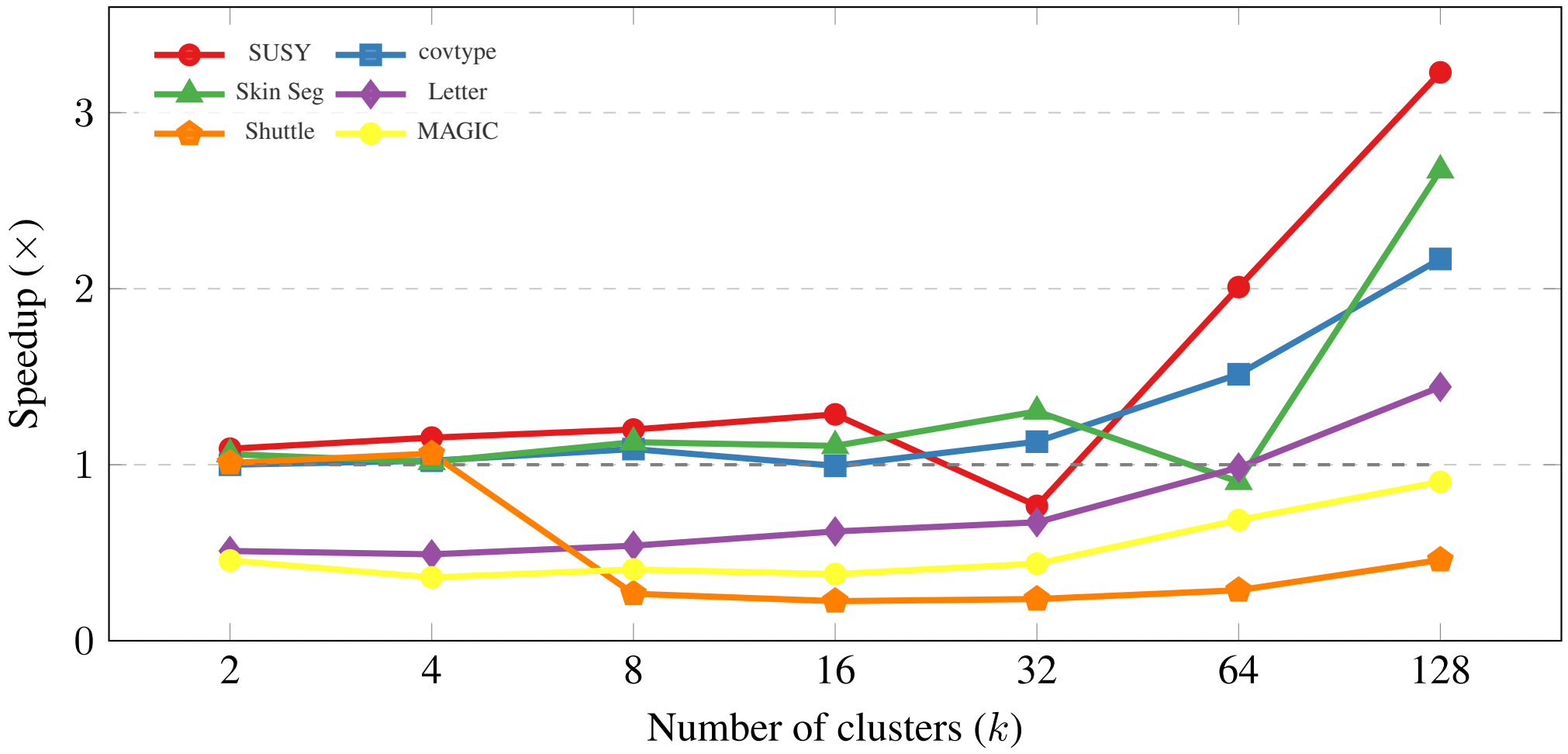

\begin{figure*}[t]
\centering
\begin{minipage}[t]{0.48\textwidth}
\begin{algorithm}[H]
\caption{Baseline GPU Batch Processing}
\label{alg:baseline}
\small
\begin{algorithmic}
\State \textbf{Input:} Batch $X_b \in \mathbb{R}^{B \times D}$, centers $C \in \mathbb{R}^{k \times D}$
\State \textbf{GPU:} Transfer $X_b$, $C$ to GPU
\State \textbf{GPU:} Compute $\hat{D} = \text{GEMM}(X_b, C)$
\State \textbf{Transfer:} $\hat{D} \rightarrow$ CPU
\State \textbf{CPU:} Compute tentative assignments
\For{$i = 1$ to $B$}
    \For{$j = 1$ to $k$}
        \State Compute weighted distance
    \EndFor
    \State $j^* \leftarrow \arg\min$ weighted distance
\EndFor
\State \textbf{CPU:} S1/S2 safety checks
\State \textbf{CPU:} Update centers, cluster sizes
\State \textbf{Output:} Updated centers, labels
\end{algorithmic}
\end{algorithm}
\end{minipage}
\hfill
\begin{minipage}[t]{0.48\textwidth}
\begin{algorithm}[H]
\caption{Optimized GPU Batch Processing}
\label{alg:optimized}
\small
\begin{algorithmic}
\State \textbf{Input:} Batch $X_b \in \mathbb{R}^{B \times D}$, \textcolor{blue}{persistent $C_{gpu} \in \mathbb{R}^{k \times D}$}
\State \textbf{GPU:} Transfer $X_b$ to GPU only
\State \textcolor{blue}{\textbf{GPU:} Compute $\hat{D}_{gpu} = \text{GEMM}(X_b, C_{gpu})$ [keep on GPU]}
\State \textcolor{blue}{\textbf{GPU:} Vectorized tentative assignments}
\State \textcolor{blue}{$W \leftarrow$ broadcast weights $(B \times k)$}
\State \textcolor{blue}{$\text{costs} \leftarrow \hat{D}_{gpu} \odot W$}
\State \textcolor{blue}{$j^* \leftarrow \arg\min(\text{costs}, \text{axis}=1)$}
\State \textcolor{blue}{\textbf{GPU:} S1/S2 safety checks (vectorized)}
\State \textbf{Transfer:} Safety mask $\in \{0,1\}^B \rightarrow$ CPU
\State \textbf{CPU:} Update centers, cluster sizes
\State \textbf{Output:} Updated centers, labels
\end{algorithmic}
\end{algorithm}
\end{minipage}
\caption{Comparison of baseline and optimized GPU batch processing.
\textbf{Key changes (blue):}
(1) Keep distance matrix $\hat{D}$ on GPU throughout processing;
(2) Vectorize tentative decisions as parallel matrix operations instead of nested CPU loops;
(3) Compute safety checks on GPU, transfer only boolean mask;
(4) Persist centers on GPU across batches (not shown).
Data transfer reduced from $O(Bk)$ floats to $O(B)$ bools.
Speedup scales with $k$ because GPU parallelism increases while fixed overheads remain constant.}
\label{fig:algorithm_comparison}
\end{figure*}

%% file: checklist.tex
\section*{NeurIPS Paper Checklist}

\begin{enumerate}

\item {\bf Claims}
    \item[] Question: Do the main claims made in the abstract and introduction accurately reflect the paper's contributions and scope?
    \item[] Answer: \answerYes{}
    \item[] Justification: The abstract and introduction state the computational motivation, the Batched Hartigan contribution, and the main empirical claims. These claims are supported by the algorithmic development in Sections~\ref{sec:fast_hartigan}--\ref{sec:results} and the appendix experiments.

\item {\bf Limitations}
    \item[] Question: Does the paper discuss the limitations of the work performed by the authors?
    \item[] Answer: \answerYes{}
    \item[] Justification: The paper discusses important limitations in Sections~\ref{sec:adaptive}, \ref{sec:runtime}, and \ref{sec:conclusion}, including the dependence on batch size, dimensionality, and the limited current benefit for multi-core/GPU-style parallelism.

\item {\bf Theory assumptions and proofs}
    \item[] Question: For each theoretical result, does the paper provide the full set of assumptions and a complete (and correct) proof?
    \item[] Answer: \answerYes{}
    \item[] Justification: The theoretical statements define their assumptions in Sections~\ref{sec:safe-reassign}--\ref{sec:pruning}. Complete proofs for the stated lemmas and theorems are provided in Appendix~\ref{app:proofs}.

    \item {\bf Experimental result reproducibility}
    \item[] Question: Does the paper fully disclose all the information needed to reproduce the main experimental results of the paper to the extent that it affects the main claims and/or conclusions of the paper (regardless of whether the code and data are provided or not)?
    \item[] Answer: \answerYes{}
    \item[] Justification: The paper gives the algorithms, datasets, tested $k$ values, number of trials, initialization, and hardware in Section~\ref{sec:results} and Appendix~\ref{app:ablation}.

\item {\bf Open access to data and code}
    \item[] Question: Does the paper provide open access to the data and code, with sufficient instructions to faithfully reproduce the main experimental results, as described in supplemental material?
    \item[] Answer: \answerNo{}
    \item[] Justification: The code will be released soon; the bureaucracy is expanding to meet the needs of the expanding bureaucracy. Sorry about that.

\item {\bf Experimental setting/details}
    \item[] Question: Does the paper specify all the training and test details (e.g., data splits, hyperparameters, how they were chosen, type of optimizer) necessary to understand the results?
    \item[] Answer: \answerYes{}
    \item[] Justification: Section~\ref{sec:results} specifies the initialization, tested $k$ range, trial count, aggregation statistic, compared algorithms, dataset scale, and single-threaded hardware setting. Appendix~\ref{app:ablation} gives additional algorithm hyperparameters and ablation settings.

\item {\bf Experiment statistical significance}
    \item[] Question: Does the paper report error bars suitably and correctly defined or other appropriate information about the statistical significance of the experiments?
    \item[] Answer: \answerNo{}
    \item[] Justification: The main results report medians over independent trials but do not provide error bars, confidence intervals, or significance tests.

\item {\bf Experiments compute resources}
    \item[] Question: For each experiment, does the paper provide sufficient information on the computer resources (type of compute workers, memory, time of execution) needed to reproduce the experiments?
    \item[] Answer: \answerYes{}
    \item[] Justification: Section~\ref{sec:results} reports single-threaded wall-clock times on an Apple M4 Pro. No exotic compute is needed. 
    
\item {\bf Code of ethics}
    \item[] Question: Does the research conducted in the paper conform, in every respect, with the NeurIPS Code of Ethics \url{https://neurips.cc/public/EthicsGuidelines}?
    \item[] Answer: \answerYes{}
    \item[] Justification: The work studies a clustering algorithm and evaluates it on benchmark datasets without human-subject experiments, high-risk model release, or privacy-sensitive data collection described in the paper. We are not aware of any aspect of the research that deviates from the NeurIPS Code of Ethics.

\item {\bf Broader impacts}
    \item[] Question: Does the paper discuss both potential positive societal impacts and negative societal impacts of the work performed?
    \item[] Answer: \answerNo{}
    \item[] Justification: We don't think anyone is concerned that clustering could lead to evil AGI.

\item {\bf Safeguards}
    \item[] Question: Does the paper describe safeguards that have been put in place for responsible release of data or models that have a high risk for misuse (e.g., pre-trained language models, image generators, or scraped datasets)?
    \item[] Answer: \answerNA{}
    \item[] Justification: The paper does not release high-risk pretrained models, image generators, scraped datasets, or other assets with a high risk of misuse. Therefore special release safeguards are not applicable.

\item {\bf Licenses for existing assets}
    \item[] Question: Are the creators or original owners of assets (e.g., code, data, models), used in the paper, properly credited and are the license and terms of use explicitly mentioned and properly respected?
    \item[] Answer: \answerNo{}
    \item[] Justification: The paper cites the algorithms and implementations used for comparison, but it does not explicitly state dataset/code package versions, licenses, or terms of use for all existing assets. 

\item {\bf New assets}
    \item[] Question: Are new assets introduced in the paper well documented and is the documentation provided alongside the assets?
    \item[] Answer: \answerNA{}
    \item[] Justification: Released code will be licensed at its Github repo.

\item {\bf Crowdsourcing and research with human subjects}
    \item[] Question: For crowdsourcing experiments and research with human subjects, does the paper include the full text of instructions given to participants and screenshots, if applicable, as well as details about compensation (if any)? 
    \item[] Answer: \answerNA{}
    \item[] Justification: The paper does not involve crowdsourcing experiments or research with human subjects. No participant instructions, screenshots, or compensation details are applicable.

\item {\bf Institutional review board (IRB) approvals or equivalent for research with human subjects}
    \item[] Question: Does the paper describe potential risks incurred by study participants, whether such risks were disclosed to the subjects, and whether Institutional Review Board (IRB) approvals (or an equivalent approval/review based on the requirements of your country or institution) were obtained?
    \item[] Answer: \answerNA{}
    \item[] Justification: The paper does not involve crowdsourcing or research with human subjects. IRB or equivalent approval details are therefore not applicable.

\item {\bf Declaration of LLM usage}
    \item[] Question: Does the paper describe the usage of LLMs if it is an important, original, or non-standard component of the core methods in this research? Note that if the LLM is used only for writing, editing, or formatting purposes and does \emph{not} impact the core methodology, scientific rigor, or originality of the research, declaration is not required.
    \item[] Answer: \answerNA{}
    \item[] Justification: The core method development does not use LLMs as an important, original, or non-standard research component. Any LLM use limited to writing, editing, or formatting would not require declaration under this checklist item.

\end{enumerate}